\documentclass[letterpaper]{article} 
\usepackage[draft]{aaai2026}  
\usepackage{times}  
\usepackage{helvet}  
\usepackage{courier}  
\usepackage[hyphens]{url}  
\usepackage{graphicx} 
\usepackage{natbib}  
\usepackage{caption} 
\usepackage{algorithm}

\usepackage{microtype}
\usepackage{amsthm}
\usepackage{booktabs}
\usepackage{amsmath}
\usepackage{amsfonts}
\usepackage{bm}
\usepackage{enumitem}
\usepackage{algpseudocode}
\makeatletter
\renewcommand*\env@matrix[1][*\c@MaxMatrixCols c]{%
  \hskip -\arraycolsep
  \let\@ifnextchar\new@ifnextchar
  \array{#1}}
\makeatother
\usepackage{newfloat}
\usepackage{listings}
\DeclareCaptionStyle{ruled}{labelfont=normalfont,labelsep=colon,strut=off} 
\floatstyle{ruled}
\newfloat{listing}{tb}{lst}{}
\floatname{listing}{Listing}
\usepackage{microtype}
\usepackage{amsthm}
\usepackage{booktabs}
\usepackage{amsmath}
\usepackage{amsfonts}
\usepackage{bm}
\usepackage{enumitem}
\DeclareMathAlphabet\mathbfcal{OMS}{cmsy}{b}{n}
\newcommand{\ten}[1]{\mathbfcal{#1}}
\newcommand{\mat}[1]{\mathbf{#1}}

\newtheorem{proposition}{Proposition}

\newcommand{\method}{\textbf{Khan-GCL}} 
\newcommand{\CKFI}{\textbf{CKFI}} 

\title{Khan-GCL: Kolmogorov–Arnold Network Based Graph Contrastive Learning with Hard Negatives}
\author{
    Zihu Wang, Boxun Xu, Hejia Geng, Peng Li\\
}
\affiliations{
    University of California, Santa Barbara\\

    \{zihu\_wang, boxunxu, hejia, lip\}@ucsb.edu
}

\usepackage{bibentry}

\begin{document}

\maketitle

\begin{abstract}
 Graph contrastive learning (GCL) has demonstrated great promise for learning generalizable graph representations from unlabeled data. However, conventional GCL approaches face two critical limitations: (1) the restricted expressive capacity of multilayer perceptron (MLP) based encoders, and (2) suboptimal negative samples that are either generated from random augmentations—failing to provide effective `hard negatives'—or hard negatives crafted without addressing the semantic distinctions crucial for discriminating graph data. To this end, we propose \method, a novel framework that integrates the Kolmogorov–Arnold Network (KAN) into the GCL encoder architecture, substantially enhancing its representational capacity. Furthermore, we exploit the rich information embedded within KAN coefficient parameters to develop two novel critical feature identification techniques that enable the generation of semantically meaningful hard negative samples for each graph representation. These strategically constructed hard negatives guide the encoder to learn more discriminative features by emphasizing critical semantic differences between graphs. Extensive experiments demonstrate that our approach achieves state-of-the-art performance compared to existing GCL methods across a variety of datasets and tasks.
\end{abstract}


\section{Introduction}

Graph Neural Networks (GNNs) are a class of machine learning models designed to learn from graph-structured data and are critical for tasks such as social network analysis, molecular property prediction, and recommendation systems. Integrating self-supervised contrastive learning (CL) that has gained popularity across a variety of domains \cite{oord2018representation, chen2020simple, gao2021simcse, radford2021learning, wang2024learn, wang2023recognizing, wang2023contrastive, wang2024learning} into graph learning has given rise to Graph Contrastive Learning (GCL), enabling pre-training GNN encoders from unlabeled graph data~\cite{velivckovic2018deep, you2020graph, zhu2020deep, you2021graph}. 

However, how to train good GCL models for real-world applications where labeled graphs are unavailable is confronted with two challenges. First, existing GCL models employ Multilayer Perceptron (MLP)-based encoders while facing a dilemma: shallow MLPs limit the generalization ability of the encoder~\cite{zhang2024expressive} while deep MLPs can easily overfit~\cite{chen2022bag,rong2019dropedge}. In addition, deep GNN encoders can overcompress, distort, or homogenize node features, making node representations indistinguishable from each other and leading to performance degradations~\cite{li2018deeper}. These difficulties have rendered use of MLP encoders with a limited depth. But in general, while being critical, striking a good balance between expressiveness and need for mitigating deep GNNs' inherent limitations is difficult.  

Second, the performance of GCL heavily relies on the construction of augmented graph data pairs. Positive pairs consist of views derived from the same graph using augmentations \cite{you2020graph,you2021graph}, which help the encoder learn semantically similar graph features. Conversely, negative pairs, comprising different graphs, provide crucial information about semantic differences and thus encourage the learning of discriminative features \cite{you2020graph,you2021graph,xia2021progcl}. In the CL literature, \textit{hard negatives} refer to samples from different classes that share similar latent semantic features with a target data point. Recent studies \cite{kalantidis2020hard,xia2021progcl,luo2023self,wang2024afans} demonstrate that incorporating such hard negatives significantly improves the encoder's performance on downstream tasks by making contrastive loss minimization more challenging. However, generating high-quality hard negatives remains a non-trivial task. 
The methods in \cite{chen2020simple,cui2021evaluating} enlarge the training batch size to include more negatives, but without guaranteeing inclusion of more hard negatives, this can lead to performance degradation\cite{kalantidis2020hard}. Additionally, the absence of labels in unsupervised pre-training renders the introduction of `false negatives', formed by pairs of samples belonging to the same class \cite{kalantidis2020hard,xia2021progcl}. Adversarial approaches generate negatives without explicitly identifying which latent features are most crucial to discriminate negative pairs \cite{hu2021adco,luo2023self,zhang2024adaptive,wang2024afans}. Therefore, more effective methods are desired for generating hard negative pairs to improve the performance of GCL.

\begin{figure*}[ht]
  \centering
  \includegraphics[width=0.85\textwidth]{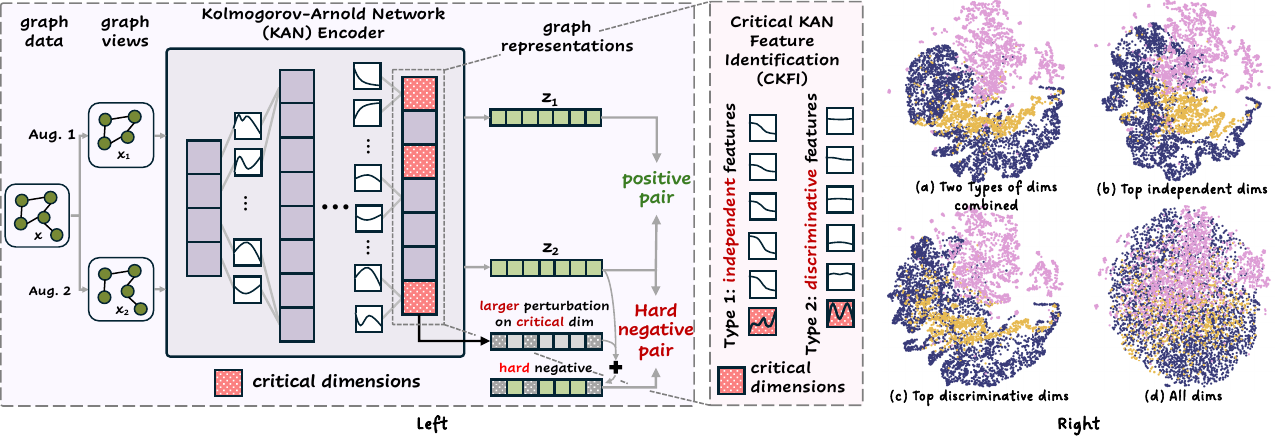}
  \caption{\textbf{Left}: Overview of the Khan-GCL framework. The encoder utilizes KAN to enhance expressive power and interpretability. Leveraging the KAN architecture, we introduce two critical dimension identification techniques. By applying small perturbations to these identified dimensions, we generate hard negative samples for each graph, thereby improving the performance of this GCL approach. \textbf{Right}: UMAP\shortcite{mcinnes2018umap} visualization of the pre-trained KAN encoder’s output feature vectors on the COLLAB dataset\cite{morris2020tudataset}. Points are colored by class (three classes). We use CKFI to calculate features' discriminative and independent scores using Equation~\ref{eq:delta} and~\ref{eq:rho}. In (d), all dimensions yield poor separation. In (b) and (c), removing less critical dimensions, the top 25\% most independent and discriminative features improve data separation. In (a), combining both types of feature dimensions (removing duplicates) achieves the best separation.}
  \label{fig:overview}
\end{figure*}

We believe that tackling the challenges brought by lack of labeled graph data and the inherent GNN problems in real-world applications requires advances in both GCL model architecture and data augmentation. To this end,  we propose \textbf{Khan-GCL}: \textbf{\underline{K}}AN-based \textbf{\underline{ha}}rd \textbf{\underline{n}}egative generation for \textbf{\underline{GCL}}. In terms of model architecture, we replace typical MLP encoders with Kolmogorov-Arnold Network (KAN) \cite{liu2024kan}. 
By integrating the Kolmogorov-Arnold representation theorem into modern neural networks, KANs introduce parameterized and learnable non-linear activation functions in the network, replacing traditional fixed activation functions in MLPs \cite{hornik1989multilayer,cybenko1989approximation}. 
KANs impose a localized structure on the trainable functions through their spline-based kernel activations. This acts as a form of regularization, guiding the model to learn smoother mappings that generalize better. This increases representational power without requiring a deeper network, which is critical for avoiding overfitting when data is scarce. As a result, KAN-based encoders strike a better balance between expensiveness and risks of inherent issues in deep GNNs.

In terms of data augmentation, we develop a method, called Critical KAN Feature Identification (\CKFI), for generating hard negatives in the output representation space of KAN encoders. By exploiting the nature of the B-spline coefficients, \CKFI\space identifies two types of critical features—\emph{discriminative} and \emph{independent} features, highlighting the most sensitive and distinctive dimensions of the underlying graph structure. Applying small perturbations to these critical features changes the essential semantics of the graph while keeping it similar to the original graph, hence generates hard negative pairs. Such high-quality hard negatives improve the encoder's ability to discriminate subtle but crucial semantics in downstream tasks.

Our main contributions of this paper include:
\begin{enumerate}[%
  label=\arabic*.,       
  labelwidth=1em,        
  labelsep=0.5em,          
  leftmargin=!           
]
   
    \item We propose \method, the first graph contrastive learning framework that integrates Kolmogorov-Arnold Network (KAN) encoders into contrastive learning, increasing the expressive power of GNNs.
    \item We introduce \CKFI, a novel approach for identifying two types of critical KAN output features, which are most independent and most discriminative of varying underlying graph structures, by exploiting the global nature of the learned B-spline coefficients from KAN encoders.
    \item We present a new method that minimally perturbs the most critical output features of each KAN encoder to generate semantically meaningful hard negatives, thereby enhancing the effectiveness of graph contrastive learning.
\end{enumerate}

Experiments across various biochemical and social media datasets demonstrate that our method achieves state-of-the-art performance on different tasks.

\begin{figure*}[ht]
  \centering
  \includegraphics[width=0.85\textwidth]{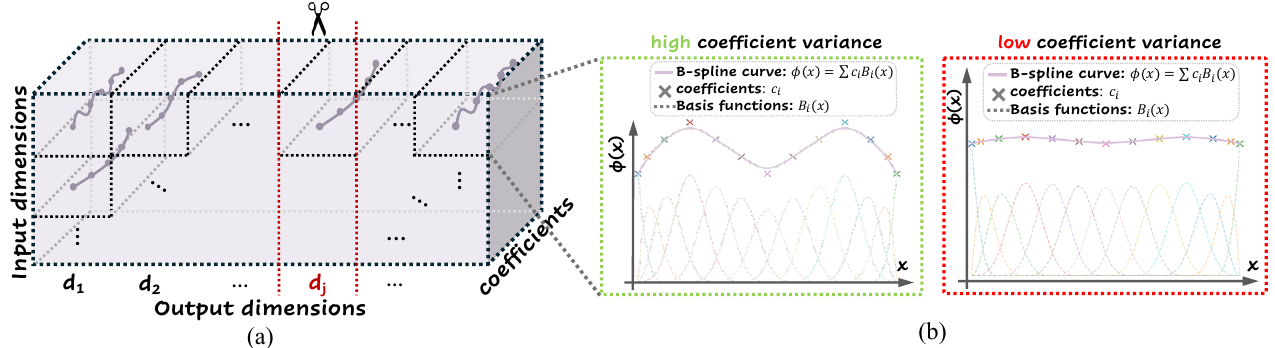}
  \caption{Illustration of a KAN layer and our two proposed critical feature identification techniques. (a) An output dimension is deemed \textit{independent} when its removal from the coefficient tensor prevents accurate reconstruction of the original tensor (i.e., results in great reconstruction error). (b) Output dimensions comprising B-splines with high coefficient variance are considered \textit{discriminative}, as larger coefficient variance typically corresponds to greater functional variance.}
  \label{fig:feature_identify}
\end{figure*}

\section{Related works}
\subsection{Graph Contrastive Learning (GCL)}

Graph Contrastive Learning (GCL) is capable of learning powerful representations from unlabeled graph data \cite{velivckovic2018deep, you2020graph, zhu2020deep, you2021graph}. Typically, GCL employs random data augmentation strategies to generate diverse views of graphs to form positive and negative pairs\cite{you2020graph}. Subsequent research has introduced automated \cite{you2021graph}, domain-knowledge informed \cite{zhu2021graph, wang2021multi}, and saliency-guided \cite{li2025self, liu2021pre, RGCL} augmentation approaches to further improve representation quality.

Recent studies in CL \cite{kalantidis2020hard, xia2021progcl} emphasize the significance of hard negatives, demonstrating their effectiveness in enhancing downstream task performance. While prior research \cite{chen2020simple,cui2021evaluating} suggests that enlarging training batch size to include more negative samples can improve feature discrimination, merely increasing the number of negatives does not inherently yield harder negative samples. In fact, continually increasing batch size can lead to performance degradation in contrastive learning \cite{kalantidis2020hard}.
To explicitly introduce hard negatives, \cite{kalantidis2020hard,xia2021progcl} propose ranking existing negatives in a mini-batch based on their similarity and mixing the hardest examples to produce hard negatives. However, the absence of labels in unsupervised pre-training may introduce false negatives. More recent approaches frame hard negative generation in GCL as an adversarial process \cite{hu2021adco,luo2023self,zhang2024adaptive,wang2024afans}, wherein a negative generator attempts to maximize the contrastive loss while the encoder aims to minimize it. However, current adversarial approaches primarily focus on bringing negative pairs closer without explicitly identifying critical latent dimensions responsible for semantic differences.

\subsection{Kolmogorov-Arnold Networks in Graph Learning}
Kolmogorov-Arnold Network (KAN)~\cite{liu2024kan} is a novel neural network architecture demonstrating improved generalization ability and interpretability over MLPs. Many recent studies~\cite{kiamari2024gkangraphkolmogorovarnoldnetworks,zhang2024graphkanenhancingfeatureextraction,bresson2025kagnnskolmogorovarnoldnetworksmeet,li2024gnnskanharnessingpowerswallowkan} have proposed KAN-based GNN architectures by directly replacing MLPs in conventional models with KAN. These studies reveal that KAN effectively enhances the expressiveness of traditional GNNs while mitigating the over-squashing problem inherent in MLP-based architectures. Furthermore, several studies have demonstrated KAN's superior generalizability over MLPs in specialized domains, including drug discovery~\cite{ahmed24graphkan}, molecular property prediction~\cite{li2024kagnnkolmogorovarnoldgraphneural}, recommendation systems~\cite{xu2024fourierkangcffourierkolmogorovarnoldnetwork}, and smart grid intrusion detection~\cite{cite-key}. 

Beyond straightforward MLP substitution, KAA \cite{fang2025kaakolmogorovarnoldattentionenhancing} embeds KANs into the attention scoring functions of GAT and Transformer-based models. KA-GAT \cite{chen2025kagat} combines KAN-based feature decomposition with multi-head attention and graph convolutions, enhancing the model’s capacity in high-dimensional graph data.

\section{Preliminary}
\paragraph{Graph Contrastive Learning} aims to train an encoder $f(\cdot)$ which maps graph data $\mat{x}\in\mathbb{R}^m$ to representations $\mat{z}\in\mathbb{R}^n$. In pre-training, a projection head $h(\cdot)$ is often employed to project the representations to $v\in\mathbb{R}^p$. GCL typically applies two random augmentation functions $A_1(\cdot)$ and $A_2(\cdot)$, sampled from a set $\mathcal{A}$ of augmentations, to produce two views of each graph from a batch $\mathcal{B}_{o}=\{\mat{x}_i^{o}\}_{i=1}^{N}$ to get a batch of augmented graphs $\mathcal{B}=\{\mat{x}_i\}_{i=1}^{2N}$, where $\mat{x}_{2i}=A_1(\mat{x}_i^0)$, $\mat{x}_{2i+1}=A_2(\mat{x}_i^0)$. These views are then encoded and projected to $\mathbb{R}^p$, i.e., $\mat{z}_i=f(\mat{x}_i)$ and $\mat{v}_i=h(\mat{z}_i)$. A contrastive loss~\cite{chen2020simple,you2020graph} is applied to encourage similarity between positive pairs and dissimilarity between negative pairs:

{
  \fontsize{8pt}{8pt}\selectfont
\begin{equation}
\begin{split}
    \mathcal{L}_{CL}=\frac{1}{2N}&\sum_{i=1}^N{-}[\mathrm{log}\frac{\mathrm{exp}(\mathrm{sim}(\mat{v}_{2i},\mat{v}_{2i+1})/\tau)}{\sum_{j\neq2i}\mathrm{exp}(\mathrm{sim}(\mat{v}_{2i},\mat{v}_j)/\tau)} \\
    &{+}\mathrm{log}\frac{\mathrm{exp}(\mathrm{sim}(\mat{v}_{2i+1},\mat{v}_{2i})/\tau)}{\sum_{j\neq2i+1}\mathrm{exp}(\mathrm{sim}(\mat{v}_{2i+1},\mat{v}_j)/\tau)}]
\end{split}
\end{equation}}

Here $\mathrm{sim}(\cdot, \cdot)$ calculates the cosine similarity between two vectors. $\tau$ is a temperature hyperparameter. Although various GCL methods have been developed, their contrastive losses are defined similarly.

\paragraph{Kolmogorov-Arnold Networks} integrate the Kolmogorov-Arnold representation theorem into modern neural networks. A KAN layer $\mathbf{\Phi}$, which maps input data from $\mathbb{R}^{d_{in}}$ to $\mathbb{R}^{d_{out}}$, is defined as:
{\fontsize{8pt}{8pt}\selectfont
\begin{equation}
    x_j^{out} = \sum_{i=1}^{d_{in}} \phi_{i,j}(x_{i}^{in}) \qquad  \forall j \in\{1,2,\dots,d_{out}\}
\label{eq:kan_in_out}
\end{equation}}
$x_{i}^{in}$ and $x_j^{out}$ denote the input and output components at dimensions $i$ and $j$, respectively. Each univariate function $\phi_{i,j}$ represents a learnable non-linear function associated with the connection from the $i_{th}$ input dimension to the $j_{th}$ output dimension. These functions are usually parameterized using B-splines~\cite{liu2024kan}, such that:
{\fontsize{8pt}{8pt}\selectfont
\begin{equation}
    \phi_{i,j}(\cdot) = \sum_k c_{ijk} B_{ijk}(\cdot)
\label{eq:bspline_function}
\end{equation}}
$B_{ijk}(\cdot)$ denotes the B-spline basis functions for $\phi_{i,j}(\cdot)$, and $c_{ijk}$ represents their corresponding trainable coefficients. All coefficients at a KAN layer can thus be denoted as $\ten{C}=\{c_{ijk}\}\in \mathbb{R}^{d_{in}\times d_{out}\times d_{c}}$, where $d_c$ is the number of coefficients used to define each B-spline function.

\section{Method}
\subsection{Overview}

Figure~\ref{fig:overview} (left) illustrates \method, the first GCL framework with a KAN-based encoder. By parameterizing non-linearity with trainable basis function coefficients, KANs offer improved generalizability and interpretability over conventional MLPs. Leveraging the rich non-linear information in KAN coefficients, we introduce \CKFI\space(Section~\ref{subsec:feature_identification}) to identify two types of critical features. Small perturbations applied to these features generate hard negatives that alter sample semantics while preserving structural similarity (Section~\ref{subsec:hardneg_generation}). Incorporating these hard negatives during pre-training guides the encoder to learn more discriminative graph features.

\subsection{Critical KAN Feature Identification (\CKFI)}
\label{subsec:feature_identification}

\subsubsection{Independent Dimensions in KANs}

According to Equation~\ref{eq:kan_in_out}, each latent dimension in a KAN layer combines a unique set of non-linear functions, capturing distinct non-linear features from the input. However, as each B-spline function is a linear combination of basis functions $B_{ijk}(\cdot)$ (Equation~\ref{eq:bspline_function}), 
any output dimension whose coefficients are linearly dependent on those of other dimensions can be expressed as a linear combination of them, indicating redundancy.

\begin{proposition}[KAN layer output dependency] 
For a KAN layer with coefficients $\ten{C}=\{c_{ijk}\}\in \mathbb{R}^{d_{in}\times d_{out}\times d_{c}}$, we denote the slice corresponding to output dimension $d$ by $\ten{C}_{:,d,:}\in \mathbb{R}^{d_{in}\times d_{c}}$. If $\ten{C}_{:,d,:}$ is a linear combination of $\ten{C}_{:,d_1,:},\ten{C}_{:,d_2,:},\dots,\ten{C}_{:,d_n,:}$, the dimension $d$'s output feature can be expressed as a linear combination of dimensions $d_1,d_2,\dots,d_n$.
\end{proposition} 
We provide the derivation of this dependency in Appendix A. Thus, conversely, we define an \textit{independent dimension} as one whose coefficients cannot be expressed as linear combinations of coefficients from other dimensions, thus encoding truly unique features globally from the entire input domain.

To identify independent dimensions in a KAN layer, as illustrated in Figure~\ref{fig:feature_identify}(a), we attempt to reconstruct the original coefficient tensor $\ten{C}$ of the layer after removing coefficients from each output dimension. Specifically, given the coefficients $\ten{C}=[c_{ijk}]\in \mathbb{R}^{d_{in}\times d_{out}\times d_{c}}$ from a KAN layer, we perform higher-order singular value decomposition (HOSVD) with each output dimension $d_j,1\leq j\leq d_{out}$ removed as follows:
{\fontsize{8pt}{10pt}\selectfont
\begin{equation}
    \ten{C}^{(-j)}\approx \ten{G}\times_1\mat{U}^{(1)}\times_2\mat{U}^{(2)}\times_3\mat{U}^{(3)}
\end{equation}}
Here $\ten{C}^{(-j)}\in \mathbb{R}^{d_{in}\times(d_{out}-1)\times d_{c}}$ denotes the tensor obtained by removing the $j_{th}$ mode-$2$ slice from $\ten{C}$. $\ten{G}\in\mathbb{R}^{r_1\times r_2\times r_3}$ is the core tensor containing singular values of $\ten{C}^{(-j)}$, and $\mat{U}^{(1)},\mat{U}^{(2)},\mat{U}^{(3)}$ are orthogonal bases for each mode of $\ten{C}^{(-j)}$. The notation $\times_n$ denotes the mode-$n$ tensor product. A detailed implementation of HOSVD is provided in Appendix B. 

Then we reconstruct the coefficients $\ten{C}$ using $\ten{C}^{(-j)}$. First, we project the removed $j_{th}$ mode-$2$ slice back to the subspace spanned by $\mat{U}^{(1)}$ and $\mat{U}^{(3)}$ to get  $\tilde{\mat{M}}_j^{(2)}$. Integrating $\tilde{\mat{M}}_j^{(2)}$ into $\mathbf{U}^{(2)}\in\mathbb{R}^{r_2\times (d_{out}-1)}$, we get $\tilde{\mathbf{U}}_j^{(2)}\in\mathbb{R}^{r_2\times d_{out}}$, the approximated basis $\mat{U}_j^{(2)}$ including the $j_{th}$ dimension. Subsequently, we reconstruct $\ten{C}$ as follows:
{\fontsize{8pt}{10pt}\selectfont
\begin{equation}  \tilde{\ten{C}}_j=\ten{G}\times_1\mat{U}^{(1)}\times_2\tilde{\mat{U}}_j^{(2)}\times_3\mat{U}^{(3)}
\end{equation}}
where $\tilde{\ten{C}}_j$ is the reconstruction with the $j_{th}$ mode-$2$ slice removed from $\ten{C}$. The reconstruction error can be calculated using the Frobenius norm:
{\fontsize{8pt}{10pt}\selectfont
\begin{equation}
    \delta_j=\Vert\tilde{\ten{C}}_j-\ten{C}\Vert_F 
    \label{eq:delta}
\end{equation}}
A larger reconstruction error $\delta_j$ indicates that the $j_{th}$ output dimension encodes more unique features, as it becomes difficult to be accurately reconstructed when excluded from the tensor decomposition. For clarity, the algorithm for identifying independent features is provided in Appendix C.

\subsubsection{Discriminative Dimensions in KANs}
In this section, we focus on \textit{discriminative dimensions}, providing an orthogonal perspective to the aforementioned independent dimensions. In downstream discrimination tasks, latent dimensions exhibiting larger output variance are preferred since they effectively separate different data points. 

To implement discriminative feature identification in practice, one can naively sample mini-batches from a dataset and calculate the variance at each dimension. However, estimating variance from randomly sampled mini-batches can introduce bias and additional computational overhead. Instead, we examine the coefficients of KAN layers, which provide a `global' view of the distribution of the underlying features. As illustrated in Figure~\ref{fig:feature_identify}(b), the shape of a B-spline curve closely aligns with the pattern of its coefficients. To formally establish this observation, we present the following proposition, establishing an upper bound on the variance of a uniform B-spline function based on its coefficients' variance.

\begin{proposition}[Variance Upper Bound of a Uniform B-spline Function]
$\phi(x) = \sum_k c_k B_k(x)$ is a uniform B-spline function defined over the interval $[a,b]$. $\{B_k(x)\}_{k=1}^{d_c}$ denotes the set of uniform B-spline basis functions and $\{c_k\}_{k=1}^{d_c}$ are the corresponding coefficients. The variance of $\phi(x)$ over its input domain satisfies the following inequality:
{\fontsize{8pt}{10pt}\selectfont
\begin{equation}
    \mathrm{Var}[\phi(x)]=\int_a^b(\phi(x)-\mu_{\phi})^2dx \leq M(0) \cdot \sigma_c^2,
    \label{eq:coef_var_bound}
\end{equation}}
where $\mu_{\phi}=\frac{1}{b-a}\int_a^b\phi(x)dx$ is the mean of $\phi(x)$, $M(0)=\int [B_k(x)]^2dx$ is identical for all $B_k(x)$ for a uniform B-spline, and $\sigma_c^2=\frac{1}{d_c}\sum_{k=1}^{d_c}(c_k-\mu_c)^2$ represents the variance of the coefficients $\{c_k\}_{k=1}^{d_c}$ with $\mu_c=\frac{1}{d_c}\sum_{k=1}^{d_c}c_k$ being the mean.
\end{proposition}
A detailed derivation of this variance upper bound of B-splines is provided in Appendix A. 


\begin{table*}
  \small
  \centering
  \setlength{\tabcolsep}{1.8mm}{
  \begin{tabular}{l|cccccccc|c}
    \toprule
    Datasets                    &BBBP  &Tox21 &ToxCast &SIDER &ClinTox &MUV &HIV &BACE &AVG\\
    \midrule
    AttrMasking\cite{hu2019strategies} &64.3$\pm$2.8 &76.7$\pm$0.4 & 64.2$\pm$0.5 &61.0$\pm$0.7 &71.8$\pm$4.1 &74.7$\pm$1.4 &77.2$\pm$1.1 &79.3$\pm$1.6 &71.1  \\
    GraphCL\cite{you2020graph}   &69.7$\pm$0.7 &73.9$\pm$0.7 &62.4$\pm$0.6 &60.5$\pm$0.9 &76.0$\pm$2.7 &69.8$\pm$2.7 &78.5$\pm$1.2 &75.4$\pm$1.4 &70.8  \\
    JOAOv2\cite{you2021graph}    &71.4$\pm$0.9 &74.3$\pm$0.6 &63.2$\pm$0.5 &60.5$\pm$0.5 &81.0$\pm$1.6 &73.7$\pm$1.0 &77.5$\pm$1.2 &75.5$\pm$1.3 &72.1  \\
    RGCL\cite{RGCL}              &71.4$\pm$0.7 &75.2$\pm$0.3 &63.3$\pm$0.2 &61.4$\pm$0.6 &83.4$\pm$0.9 &76.7$\pm$1.0 &77.9$\pm$0.8 &76.0$\pm$0.8 &73.2  \\
    GraphACL\cite{luo2023self}   &73.3$\pm$0.5 &76.2$\pm$0.6 &64.1$\pm$0.4 & 62.6$\pm$0.6 &\textbf{85.0$\pm$1.6} & \underline{76.9$\pm$1.2} & 78.9$\pm$0.7 & 80.1$\pm$1.2 &74.6  \\
    DRGCL\cite{ji2024rethinking} &71.2$\pm$0.5 &74.7$\pm$0.5 &64.0$\pm$0.5 &61.1$\pm$0.8 &78.2$\pm$1.5 &73.8$\pm$1.1 &78.6$\pm$1.0 &78.2$\pm$1.0 &72.5  \\
    CI-GCL\cite{tan2024community} & \textbf{74.4$\pm$1.9} & \underline{77.3$\pm$0.9} & \underline{65.4$\pm$1.5} & \underline{64.7$\pm$0.3} & 80.5$\pm$1.3 & 76.5$\pm$0.9 & \underline{80.5$\pm$1.3} & \textbf{84.4$\pm$0.9} & \underline{75.4} \\
    \midrule
    Khan-GCL        & \underline{73.5$\pm$0.6}  &\textbf{78.3$\pm$0.3} &\textbf{66.3$\pm$0.3} &\textbf{65.0$\pm$0.9} & \underline{84.3$\pm$1.2} &\textbf{77.5$\pm$0.4} &\textbf{80.7$\pm$0.6} &\underline{80.9$\pm$1.0} & \textbf{75.8}\\
    \bottomrule
  \end{tabular}}
  \caption{Transfer learning performance (ROC-AUC scores in \%) for graph classification across 8 datasets. Results for benchmark methods are reported from their respective original publications.}
  \label{tab:transfer_learning}
\end{table*}

Equation~\eqref{eq:coef_var_bound} implies that the variance of the output of a B-spline function over its input domain is bounded by the variance of its coefficients. Consequently, as illustrated in Figure~\ref{fig:feature_identify}(b), we exploit the variance of coefficients in each feature dimension to assess the discriminative power of that dimension. Specifically, in a KAN layer whose coefficients are denoted by $\ten{C}=[c_{ijk}]\in \mathbb{R}^{d_{in}\times d_{out}\times d_{c}}$, we quantify the discriminative capability of an output dimension $j$ using the average variance across all B-spline functions associated with that dimension as follows:
{\fontsize{8pt}{10pt}\selectfont
\begin{equation}
    \rho_j=\frac{1}{d_{in}}\cdot\sum_{i=1}^{d_{in}}\sigma_{c_{ij}}^2
    \label{eq:rho}
\end{equation}}
where $\mu_{c_{ij}}=\frac{1}{d_c}\sum_{k=1}^{d_c}c_{ijk}$, and $\sigma_{c_{ij}}^2=\frac{1}{d_c}\sum_{k=1}^{d_c}(c_{ijk}-\mu_{c_{ij}})^2$  denoting the variance of coefficients for the B-spline function $\phi_{i,j}(\cdot)$ linking the $i^{th}$ input dimension to the $j^{th}$ output dimension. A dimension $j$ with a large $\rho_j$ is considered more discriminative w.r.t. the input data.

\subsection{Hard Negatives in Khan-GCL}
\label{subsec:hardneg_generation}
Prior research~\cite{kalantidis2020hard,xia2021progcl} suggests that in contrastive learning, the most effective hard negatives for a graph satisfy two key criteria: (a) their key identity is different from the original graph, and (b) they maintain high semantic similarity to the original graph. Therefore, to generate an effective hard negative for a graph, we aim to produce a variant with \textbf{minimal deviation from the original} while strategically \textbf{distorting its key characteristics}. As shown in Figure~\ref{fig:overview} (right), two types of CKFI dimensions include the most important features to recognize data from different classes. Applying small perturbations to these dimensions can thus effectively change the identity of a feature vector.

We propose applying small perturbations to graph representations from the encoder's last layer to generate hard negatives in the output space of the encoder. Specifically, we define these perturbations as follows:
{\fontsize{8pt}{10pt}\selectfont
\begin{equation}
\begin{aligned}
    \mathbf{p}^{\delta}&=\{p_i^{\delta}=\alpha_i\cdot u_i^{\delta}:u_i^{\delta}\sim\mathcal{N}(\epsilon_{\delta}\cdot\delta_i,\sigma^2_{\delta}),\alpha_i\sim \mathrm{Rad}\} \\
    \mathbf{p}^{\rho}&=  \{p_i^{\rho}=\alpha_i\cdot u_i^{\rho}:u_i^{\rho}\sim\mathcal{N}(\epsilon_{\rho}\cdot\rho_i,\sigma^2_{\rho}),\alpha_i\sim \mathrm{Rad}\}
\end{aligned}
\end{equation}}
Here $\epsilon_{\delta}>0$ and $\epsilon_{\rho}>0$ are hyperparameters scaling $\bm{\delta}=\{\delta_i\}$ and $\bm{\rho}=\{\rho_i\}$ computed by the proposed CKFI method for feature $i$ per Equations~(\ref{eq:delta}) and (\ref{eq:rho}), respectively. $\sigma^2_{\delta}$ and $\sigma^2_{\rho}$ represent the variance hyperparameters of the Gaussian distributions. With these perturbations, dimensions that are highly discriminative or independent (i.e., those with large $\rho_i$ or $\delta_i$ values) receive perturbations from Gaussian distributions with larger means. Since $\epsilon_{\rho}\cdot\rho_i>0$ and $\epsilon_{\delta}\cdot\delta_i>0$ for all dimensions $i$, we introduce $\alpha_i$ sampled from the Rademacher distribution $\mathrm{Rad}$ to ensure that perturbations $p_i^{\rho}$ and $p_i^{\delta}$ are approximately equally likely to be positive or negative. With these perturbations critical dimensions receive more substantial perturbations on average.

During training, with a mini-batch of $N$ graphs, the augmented graph data is denoted by $\mathcal{B}=\{\mat{x}_{j}\}_{j=1}^{2N}$ of size $2N$, and their latent representations are $\mathcal{B}_z=\{\mat{z}_{j}\}_{j=1}^{2N}$. For each representation $\mat{z}_j$ in $\mathcal{B}_z$, we sample perturbation vectors $\mat{p}_j^{\rho}$ and $\mat{p}_j^{\delta}$ and produce the hard negative of $\mathbf{z}_j$ as:
{\fontsize{8pt}{10pt}\selectfont
\begin{equation}
\mat{z}^{hard}_j=\mat{z}_j+\mat{p}_j^{\rho}+\mat{p}_j^{\delta}
\end{equation}}

The generated hard negative $\mat{z}^{hard}_j$ is then projected to $\mat{v}^{hard}_j$ by the projection head and utilized in our proposed hard negative loss:
{\fontsize{8pt}{10pt}\selectfont
\begin{equation}
    \mathcal{L}_{HN}=\frac{1}{2N}\sum_{j=1}^{2N} \mathrm{log}[\mathrm{exp}(\mathrm{sim}(\mat{v}_j,\mathrm{sg}(\mat{v}_j^{hard})))]
\end{equation}}
To prevent model collapse, we apply a stop-gradient operator $\mathrm{sg}(\cdot)$ to hard negatives $\mat{v}_i^{hard}$. Finally, we write the overall training loss $\mathcal{L}_{Khan}$ of Khan-GCL as:
{\fontsize{8pt}{10pt}\selectfont
\begin{equation}
    \mathcal{L}_{Khan} = \mathcal{L}_{CL} + \mathcal{L}_{HN}
\end{equation}}
We present the detailed algorithm flow of Khan-GCL in Appendix D.

\begin{table*}
  \small
  \centering
  \setlength{\tabcolsep}{1.5mm}{
  \begin{tabular}{l|cccccccc|c}
    \toprule
    Datasets                    & DD & MUTAG &NCI1 &PROTEINS &COLLAB &RDT-B &RDT-M5K &IMDB-B &AVG\\
    \midrule
    InfoGraph~\cite{sun2019infograph} &72.9$\pm$1.8 &89.0$\pm$1.1 &76.2$\pm$1.0 &74.4$\pm$0.3 &70.1$\pm$1.1 &82.5$\pm$1.4 &53.5$\pm$1.0 &73.0$\pm$0.9 &74.0  \\
    GraphCL\cite{you2020graph}   &78.6$\pm$0.4 &86.8$\pm$1.3 &77.9$\pm$0.4 &74.4$\pm$0.5 &71.4$\pm$1.1 &89.5$\pm$0.8 &56.0$\pm$0.3 &71.1$\pm$0.4 &75.7  \\
    JOAOv2\cite{you2021graph}    &77.4$\pm$1.1 &87.7$\pm$0.8 &78.4$\pm$0.5 &74.1$\pm$1.1 &69.3$\pm$0.3 &86.4$\pm$1.5 &56.0$\pm$0.3 &70.1$\pm$0.3 &74.9\\
    AD-GCL~\cite{suresh2021adversarial} &75.8$\pm$0.9 &88.7$\pm$1.9 &73.9$\pm$0.8 &73.3$\pm$0.5 & 72.0$\pm$0.6 &90.1$\pm$0.9 &54.3$\pm$0.3 &70.2$\pm$0.7 & 74.8 \\
    RGCL\cite{RGCL}              &78.9$\pm$0.5 &87.7$\pm$1.0 &78.1$\pm$1.1 &75.0$\pm$0.4 &71.0$\pm$0.7 &90.3$\pm$0.6 & 56.4$\pm$0.4 &71.9$\pm$0.9 &76.2\\
    DRGCL\cite{ji2024rethinking} &78.4$\pm$0.7 &89.5$\pm$0.6 &78.7$\pm$0.4 &75.2$\pm$0.6 &70.6$\pm$0.8 & \underline{90.8$\pm$0.3} &56.3$\pm$0.2 &72.0$\pm$0.5 &76.4\\
    TopoGCL\shortcite{chen2024topogcl}&79.1$\pm$0.3 &\underline{90.1$\pm$0.9} &\textbf{81.3$\pm$0.3} &\textbf{77.3$\pm$0.9} &-   
                  &90.4$\pm$0.5 &-            &\underline{74.7$\pm$0.3} & - \\
    CI-GCL\cite{tan2024community} & \underline{79.6$\pm$0.3} & 89.7$\pm$0.9 & 80.5$\pm$0.5 & 76.5$\pm$0.1 & \underline{74.4$\pm$0.6} & \underline{90.8$\pm$0.5} & \underline{56.6$\pm$0.3} & 73.8$\pm$0.8 & \underline{77.7}\\
    \midrule
    Khan-GCL(Ours)              &\textbf{80.6$\pm$0.7}       &\textbf{91.4$\pm$1.1} &\underline{80.8$\pm$0.9} &\underline{76.9$\pm$0.8} &\textbf{75.2$\pm$0.3} &\textbf{92.2$\pm$0.3} &\textbf{56.9$\pm$0.5} &\textbf{75.0$\pm$0.4} &\textbf{78.6}\\
    \bottomrule
      \end{tabular}}
    \caption{Unsupervised learning performance (accuracy in \%) for graph classification on TU-datasets. Results for benchmark methods are quoted from their original publications, except for AD-GCL and InfoGraph, which are reported from \cite{RGCL}. Average performance is calculated across all 8 datasets.}
  \label{tab:unsupervised}
\end{table*}

\begin{table*}[ht]
  \small
  \centering
  \setlength{\tabcolsep}{1.6mm}{
  \begin{tabular}{l|cccccccc}
    \toprule
    Datasets                    & DD & MUTAG &NCI1 &PROTEINS &COLLAB &RDT-B &RDT-M5K &IMDB-B\\
    \midrule
    AFANS\cite{wang2024afans}     &- &90.0$\pm$1.0 &80.4$\pm$0.5 &75.4$\pm$0.5 &\underline{74.7$\pm$0.5} &\underline{91.1$\pm$0.1} &- &-   \\
    ANGCL\cite{zhang2024adaptive} &78.8$\pm$0.9 &\textbf{92.3$\pm$0.7} &\textbf{81.0$\pm$0.3} &\underline{75.9$\pm$0.4} &72.0$\pm$0.7 &90.8$\pm$0.7 &\underline{56.5$\pm$0.3} &71.8$\pm$0.6  \\
    GraphACL\cite{luo2023self}    &\underline{79.3$\pm$0.4} &90.2$\pm$0.9 &- &75.5$\pm$0.4 &\underline{74.7$\pm$0.6} &- &- &\underline{74.3$\pm$0.7}  \\
    \midrule
    Khan-GCL(Ours)              &\textbf{80.6$\pm$0.7}       &\underline{91.4$\pm$1.1} &\underline{80.8$\pm$0.9} &\textbf{76.9$\pm$0.8} &\textbf{75.2$\pm$0.3} &\textbf{92.2$\pm$0.3} &\textbf{56.9$\pm$0.5} &\textbf{75.0$\pm$0.4} \\
    \bottomrule
      \end{tabular}}
    \caption{Unsupervised learning performance (accuracy in \%) for graph classification on TU-datasets, comparing Khan-GCL against existing hard negative integrated GCL methods. Benchmark results are from their corresponding original publications.}
  \label{tab:other_hard_neg}
\end{table*}

\section{Experiments}
In this section, we conduct comprehensive experiments across diverse datasets and tasks to demonstrate the efficacy of our approach. Furthermore, we conduct extensive ablation studies to provide deeper insights into the mechanisms underlying our proposed method. In all experimental results tables, \textbf{bold values} denote the best performance on the corresponding dataset, while \underline{underlined values} indicate the second-best performance. All experiments are run on a single NVIDIA A100 GPU.

\paragraph{Model architecture and hyperparameters.} For fair comparison, we follow the general contrastive learning hyperparameter settings of ~\cite{you2020graph}. In our KAN-based encoder implementation, we systematically replace all MLPs in the backbone architectures of \cite{you2020graph} with Cubic KAN layers (utilizing 3rd order B-spline functions) while maintaining identical input, output, and hidden dimensions. The comprehensive details regarding model architecture and hyperparameter configurations are provided in Appendix F.

\paragraph{Datasets.} We conduct experiments on Zinc-2M~\cite{sterling2015zinc}, 8 biochemical datasets from~\cite{wu2018moleculenet}, 8 diverse biochemical/social network datasets from the TU-datasets collection~\cite{morris2020tudataset}, and MNIST-superpixel~\cite{monti2017geometric}. Details about the datasets are provided in Appendix E.

\subsection{Main Results}
\paragraph{Transfer learning} is a widely adopted evaluation protocol for assessing the generalizability and transferability of representations learned by GCL methods. We pre-train our backbone encoder on the large-scale molecular dataset Zinc-2M~\cite{sterling2015zinc} using the proposed Khan-GCL framework, then finetune the pre-trained encoder on 8 biochemical datasets for graph classification tasks. Detailed training and evaluation settings are provided in Appendix E. Table~\ref{tab:transfer_learning} shows the graph classification accuracy across these datasets. 

Khan-GCL achieves the best overall results compared to all state-of-the-art methods. KAN's better generalization capability over MLPs and our introduced hard negative generation technique enhance the encoder's generalizability and ability to discriminate between subtle yet critical differences across graph structures.

\paragraph{Unsupervised learning} aims to assess a GCL pre-training method's efficacy on diverse datasets. Following established protocols in~\cite{sun2019infograph,you2020graph}, we pre-train our encoder on 8 datasets from TU-datasets~\cite{morris2020tudataset}. Subsequently, we employ an SVM classifier to evaluate the pre-trained encoder's feature representation quality. More detailed setups can be found in Appendix E.

Table~\ref{tab:unsupervised} presents a comprehensive comparison between Khan-GCL and other state-of-the-art GCL approaches in unsupervised learning experiments. Khan-GCL achieves the best performance on 6 out of 8 datasets and yields the best overall results across all methods due to the powerful KAN encoder and effective generation of hard negatives. 
Furthermore, Table~\ref{tab:other_hard_neg} compares Khan-GCL's effectiveness against existing hard negative integrated GCL methods, where our approach attains optimal results on 6 out of 8 datasets. 
\begin{figure}
    \centering
    \includegraphics[width=0.85\linewidth]{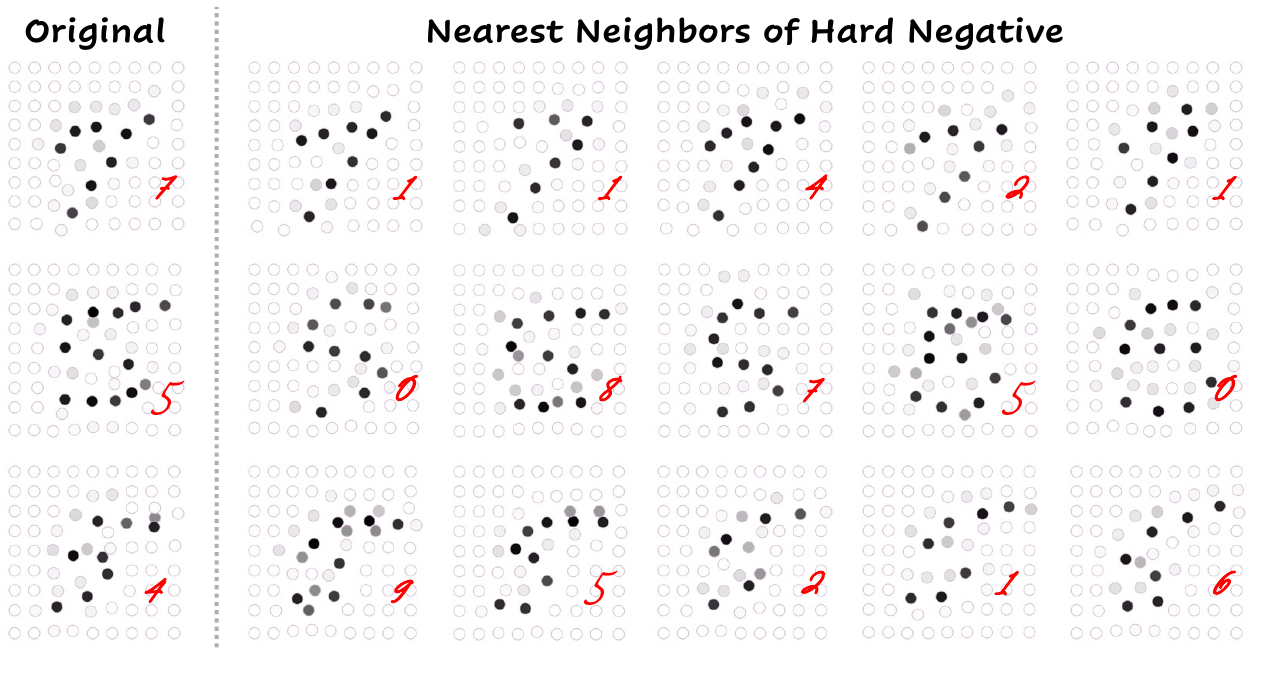}
    \caption{MNIST-superpixel graphs with their generated hard negatives' nearest neighbors in the latent space. Red numbers indicate ground truth labels. The nearest neighbors of a sample's hard negatives are typically similar to the sample while belonging to different classes.}
    \label{fig:hard_neg_retrieval}
\end{figure}

By targeting critical dimensions identified through CKFI, Khan-GCL emphasizes essential semantic features, which significantly enhance its classification capabilities compared to existing hard negative integrated methods.

\begin{figure*}[ht]
  \centering
  \includegraphics[width=0.9\textwidth]{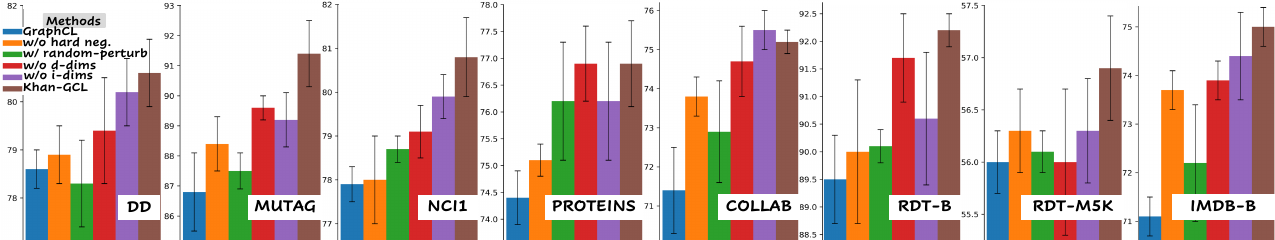}
  \caption{Ablation study results for Khan-GCL. Unsupervised learning results (in \%) for graph classification on TU-datasets are shown.}
  \label{fig:ablation}
\end{figure*}

\paragraph{Nearest neighbors retrieval of the generated hard negatives.}
To better elucidate the effectiveness of the generated hard negatives, we pre-train an encoder using Khan-GCL on MNIST-superpixel~\cite{monti2017geometric}, where handwritten digits from the MNIST dataset~\cite{lecun1998gradient} are represented as graphs. During pre-training, we generate hard negatives for each graph and retrieve the nearest neighbors of each hard negative in the feature space across the entire dataset. We provide more details of this experiment in Appendix E. Figure~\ref{fig:hard_neg_retrieval} illustrates sample digit graphs alongside the five nearest neighbors of their corresponding hard negatives. The ground truth label for each graph is displayed in red at the bottom right corner. As shown in Figure~\ref{fig:hard_neg_retrieval}, the nearest neighbors of a sample's hard negatives typically exhibit structural similarity to the original sample while belonging to different classes. This observation confirms that forming negative pairs between a sample and such generated hard negatives effectively guides the encoder to discriminate between semantically similar samples from different classes.

\begin{table} 
  \small
  \centering
  \setlength{\tabcolsep}{1mm}{
  \begin{tabular}{l|cccc}
    \toprule
    Datasets                    & DD & MUTAG &NCI1 &PROTEINS \\
    \midrule
    GraphCL &78.6$\pm$0.4 &86.8$\pm$1.3 &77.9$\pm$0.4 &74.4$\pm$0.5 \\
    GraphCL (KAN)         &\underline{78.9$\pm$0.6} &\underline{88.4$\pm$0.9} &\underline{78.0$\pm$1.0} &\underline{75.1$\pm$0.3} \\
    Ours              &\textbf{80.6$\pm$0.7}       &\textbf{91.4$\pm$1.1} &\textbf{80.8$\pm$0.9} &\textbf{76.9$\pm$0.8}\\
    \midrule
    JOAOv2      &77.4$\pm$1.1 &87.7$\pm$0.8 &78.4$\pm$0.5 &74.1$\pm$1.1 \\
    JOAOv2 (KAN)  &\underline{78.7$\pm$0.5} &\underline{88.6$\pm$0.8} &\underline{79.0$\pm$0.5} &\underline{75.3$\pm$0.6} \\
    JOAOv2+Ours              &\textbf{80.2$\pm$1.1}       &\textbf{92.1$\pm$0.3} &\textbf{81.5$\pm$0.7} &\textbf{77.0$\pm$0.9}\\
    \bottomrule
      \end{tabular}}
    \caption{Ablation study results on the effectiveness of KAN in Khan-GCL. Unsupervised learning results (in \%) for graph classification on TU-datasets are reported.}
  \label{tab:ablation_kan}
\end{table}

\subsection{Ablation Study}
\paragraph{Effectiveness of the KAN Encoder and Compatibility of Khan-GCL with Existing GCL Methods.}
We conduct targeted experiments to assess the KAN encoder’s impact on Khan-GCL. Specifically, we evaluate Khan-GCL variants without CKFI and hard negative generation (`GraphCL (KAN)' and `JOAOv2 (KAN)' in Table~\ref{tab:ablation_kan}) against their baselines. We also introduce `JOAOv2+Ours', which applies the full Khan-GCL framework to JOAOv2. Results show that KAN-enhanced variants outperform their counterparts even without hard negatives, demonstrating KAN’s superior ability in modeling non-linearity. Adding hard negatives yields further gains, confirming its effectiveness and Khan-GCL’s compatibility with diverse GCL methods.

\paragraph{Effectiveness of two feature identification techniques in hard negative generation.}
In this section, we evaluate the contributions of our two proposed critical feature identification techniques in hard negative generation. We introduce three variants of Khan-GCL: (1) `w/o d-dims', where negatives are generated by perturbing only independent dimensions; (2) `w/o i-dims', where perturbation is limited to discriminative dimensions; and (3) `w/ rand-perturb', which generates negatives by applying random Gaussian noise across all dimensions. Detailed experimental settings for these configurations are provided in Appendix E. Figure~\ref{fig:ablation} presents the performance of these configurations on unsupervised learning tasks. While `w/ rand-perturb' yields performance improvements over `w/o hard neg.' (a Khan-GCL variant with KAN encoder but without hard negative generation) on several datasets, it occasionally results in performance degradation. Both specialized perturbation approaches (`w/o d-dims' and `w/o i-dims') outperform the baseline, demonstrating the effectiveness of targeted feature identification technique. Additionally, Khan-GCL, which integrates both proposed feature identification techniques in CKFI, achieves the most substantial improvement and the best overall results, confirming the complementary nature of our dual feature identification approach.


\section{Conclusion}
We propose \method, the first KAN-based graph contrastive learning framework, which balances expressive power and risks of inherent issues of deep GNNs by using a KAN-based encoder. We also introduce \CKFI\space to identify discriminative and independent features, enabling the generation of hard negatives through minimal perturbations. These hard negatives guide the encoder to learn critical semantics during contrastive pre-training. Extensive experiments on biochemical and social network datasets demonstrate that our approach significantly improves generalization and transferability, achieving the state-of-the-art performance. Additional details and experimental results are provided in the Appendix.

For future work, exploring feature perturbation and hard negative generation in intermediate layers of a KAN encoder is promising. Further, reducing the additional training cost introduced by KAN’s spline computations through more efficient architectures is worth investigating.

\bibliography{aaai2026}

\begin{thebibliography}{58}
\providecommand{\natexlab}[1]{#1}

\bibitem[{Ahmed and Sifat(2024)}]{ahmed24graphkan}
Ahmed, T.; and Sifat, M. H.~R. 2024.
\newblock GraphKAN: Graph Kolmogorov Arnold Network for Small Molecule-Protein Interaction Predictions.
\newblock In \emph{ICML'24 Workshop ML for Life and Material Science: From Theory to Industry Applications}.

\bibitem[{Bresson et~al.(2025)Bresson, Nikolentzos, Panagopoulos, Chatzianastasis, Pang, and Vazirgiannis}]{bresson2025kagnnskolmogorovarnoldnetworksmeet}
Bresson, R.; Nikolentzos, G.; Panagopoulos, G.; Chatzianastasis, M.; Pang, J.; and Vazirgiannis, M. 2025.
\newblock KAGNNs: Kolmogorov-Arnold Networks meet Graph Learning.
\newblock arXiv:2406.18380.

\bibitem[{Chen et~al.(2025)Chen, Yuchi, Yan, Dong, and Li}]{chen2025kagat}
Chen, J.; Yuchi, X.; Yan, Z.; Dong, K.; and Li, H. 2025.
\newblock {KA}-{GAT}: Kolmogorov{\textendash}Arnold based Graph Attention Networks.

\bibitem[{Chen et~al.(2020)Chen, Kornblith, Norouzi, and Hinton}]{chen2020simple}
Chen, T.; Kornblith, S.; Norouzi, M.; and Hinton, G. 2020.
\newblock A simple framework for contrastive learning of visual representations.
\newblock In \emph{International conference on machine learning}, 1597--1607. PmLR.

\bibitem[{Chen et~al.(2022)Chen, Zhou, Duan, Zheng, Wang, Hu, and Wang}]{chen2022bag}
Chen, T.; Zhou, K.; Duan, K.; Zheng, W.; Wang, P.; Hu, X.; and Wang, Z. 2022.
\newblock Bag of tricks for training deeper graph neural networks: A comprehensive benchmark study.
\newblock \emph{IEEE Transactions on Pattern Analysis and Machine Intelligence}, 45(3): 2769--2781.

\bibitem[{Chen, Frias, and Gel(2024)}]{chen2024topogcl}
Chen, Y.; Frias, J.; and Gel, Y.~R. 2024.
\newblock TopoGCL: Topological Graph Contrastive Learning.
\newblock \emph{arXiv preprint arXiv:2406.17251}.

\bibitem[{Cui et~al.(2021)Cui, Du, Yang, Zhou, Xu, Zhou, Cheng, and Liu}]{cui2021evaluating}
Cui, G.; Du, Y.; Yang, C.; Zhou, J.; Xu, L.; Zhou, X.; Cheng, X.; and Liu, Z. 2021.
\newblock Evaluating modules in graph contrastive learning.
\newblock \emph{arXiv preprint arXiv:2106.08171}.

\bibitem[{Cybenko(1989)}]{cybenko1989approximation}
Cybenko, G. 1989.
\newblock Approximation by superpositions of a sigmoidal function.
\newblock \emph{Mathematics of control, signals and systems}, 2(4): 303--314.

\bibitem[{De~Lathauwer, De~Moor, and Vandewalle(2000)}]{de2000multilinear}
De~Lathauwer, L.; De~Moor, B.; and Vandewalle, J. 2000.
\newblock A multilinear singular value decomposition.
\newblock \emph{SIAM journal on Matrix Analysis and Applications}, 21(4): 1253--1278.

\bibitem[{Fang et~al.(2025)Fang, Gao, Wang, Shang, Chow, Chen, and Yang}]{fang2025kaakolmogorovarnoldattentionenhancing}
Fang, T.; Gao, T.; Wang, C.; Shang, Y.; Chow, W.; Chen, L.; and Yang, Y. 2025.
\newblock KAA: Kolmogorov-Arnold Attention for Enhancing Attentive Graph Neural Networks.
\newblock arXiv:2501.13456.

\bibitem[{Fey and Lenssen(2019)}]{fey2019fast}
Fey, M.; and Lenssen, J.~E. 2019.
\newblock Fast graph representation learning with PyTorch Geometric.
\newblock \emph{arXiv preprint arXiv:1903.02428}.

\bibitem[{Gao, Yao, and Chen(2021)}]{gao2021simcse}
Gao, T.; Yao, X.; and Chen, D. 2021.
\newblock Simcse: Simple contrastive learning of sentence embeddings.
\newblock \emph{arXiv preprint arXiv:2104.08821}.

\bibitem[{Hornik, Stinchcombe, and White(1989)}]{hornik1989multilayer}
Hornik, K.; Stinchcombe, M.; and White, H. 1989.
\newblock Multilayer feedforward networks are universal approximators.
\newblock \emph{Neural networks}, 2(5): 359--366.

\bibitem[{Hu et~al.(2021)Hu, Wang, Hu, and Qi}]{hu2021adco}
Hu, Q.; Wang, X.; Hu, W.; and Qi, G.-J. 2021.
\newblock Adco: Adversarial contrast for efficient learning of unsupervised representations from self-trained negative adversaries.
\newblock In \emph{Proceedings of the IEEE/CVF Conference on Computer Vision and Pattern Recognition}, 1074--1083.

\bibitem[{Hu et~al.(2019)Hu, Liu, Gomes, Zitnik, Liang, Pande, and Leskovec}]{hu2019strategies}
Hu, W.; Liu, B.; Gomes, J.; Zitnik, M.; Liang, P.; Pande, V.; and Leskovec, J. 2019.
\newblock Strategies for pre-training graph neural networks.
\newblock \emph{arXiv preprint arXiv:1905.12265}.

\bibitem[{Ji et~al.(2024)Ji, Li, Hu, Wang, Zheng, and Xu}]{ji2024rethinking}
Ji, Q.; Li, J.; Hu, J.; Wang, R.; Zheng, C.; and Xu, F. 2024.
\newblock Rethinking dimensional rationale in graph contrastive learning from causal perspective.
\newblock In \emph{Proceedings of the AAAI Conference on Artificial Intelligence}, volume~38, 12810--12820.

\bibitem[{Kalantidis et~al.(2020)Kalantidis, Sariyildiz, Pion, Weinzaepfel, and Larlus}]{kalantidis2020hard}
Kalantidis, Y.; Sariyildiz, M.~B.; Pion, N.; Weinzaepfel, P.; and Larlus, D. 2020.
\newblock Hard negative mixing for contrastive learning.
\newblock \emph{Advances in neural information processing systems}, 33: 21798--21809.

\bibitem[{Kiamari, Kiamari, and Krishnamachari(2024)}]{kiamari2024gkangraphkolmogorovarnoldnetworks}
Kiamari, M.; Kiamari, M.; and Krishnamachari, B. 2024.
\newblock GKAN: Graph Kolmogorov-Arnold Networks.
\newblock arXiv:2406.06470.

\bibitem[{Kingma(2014)}]{kingma2014adam}
Kingma, D.~P. 2014.
\newblock Adam: A method for stochastic optimization.
\newblock \emph{arXiv preprint arXiv:1412.6980}.

\bibitem[{LeCun et~al.(1998)LeCun, Bottou, Bengio, and Haffner}]{lecun1998gradient}
LeCun, Y.; Bottou, L.; Bengio, Y.; and Haffner, P. 1998.
\newblock Gradient-based learning applied to document recognition.
\newblock \emph{Proceedings of the IEEE}, 86(11): 2278--2324.

\bibitem[{Li et~al.(2024{\natexlab{a}})Li, Zhang, Wang, and Xia}]{li2024kagnnkolmogorovarnoldgraphneural}
Li, L.; Zhang, Y.; Wang, G.; and Xia, K. 2024{\natexlab{a}}.
\newblock KA-GNN: Kolmogorov-Arnold Graph Neural Networks for Molecular Property Prediction.
\newblock arXiv:2410.11323.

\bibitem[{Li, Han, and Wu(2018)}]{li2018deeper}
Li, Q.; Han, Z.; and Wu, X.-M. 2018.
\newblock Deeper insights into graph convolutional networks for semi-supervised learning.
\newblock In \emph{Proceedings of the AAAI conference on artificial intelligence}, volume~32.

\bibitem[{Li et~al.(2024{\natexlab{b}})Li, Li, Liu, and Chen}]{li2024gnnskanharnessingpowerswallowkan}
Li, R.; Li, M.; Liu, W.; and Chen, H. 2024{\natexlab{b}}.
\newblock GNN-SKAN: Harnessing the Power of SwallowKAN to Advance Molecular Representation Learning with GNNs.
\newblock arXiv:2408.01018.

\bibitem[{Li et~al.(2025)Li, Luo, Zhang, Wang, Li, Zhou, and Chua}]{li2025self}
Li, S.; Luo, Y.; Zhang, A.; Wang, X.; Li, L.; Zhou, J.; and Chua, T.-S. 2025.
\newblock Self-attentive rationalization for interpretable graph contrastive learning.
\newblock \emph{ACM Transactions on Knowledge Discovery from Data}, 19(2): 1--21.

\bibitem[{Li et~al.(2022)Li, Wang, Zhang, He, and Chua}]{RGCL}
Li, S.; Wang, X.; Zhang, A.; He, X.; and Chua, T.-S. 2022.
\newblock Let Invariant Rationale Discovery Inspire Graph Contrastive Learning.
\newblock In \emph{{ICML}}.

\bibitem[{Liu et~al.(2021)Liu, Wang, Liu, Lasenby, Guo, and Tang}]{liu2021pre}
Liu, S.; Wang, H.; Liu, W.; Lasenby, J.; Guo, H.; and Tang, J. 2021.
\newblock Pre-training molecular graph representation with 3d geometry.
\newblock \emph{arXiv preprint arXiv:2110.07728}.

\bibitem[{Liu et~al.(2024)Liu, Wang, Vaidya, Ruehle, Halverson, Solja{\v{c}}i{\'c}, Hou, and Tegmark}]{liu2024kan}
Liu, Z.; Wang, Y.; Vaidya, S.; Ruehle, F.; Halverson, J.; Solja{\v{c}}i{\'c}, M.; Hou, T.~Y.; and Tegmark, M. 2024.
\newblock Kan: Kolmogorov-arnold networks.
\newblock \emph{arXiv preprint arXiv:2404.19756}.

\bibitem[{Luo et~al.(2023)Luo, Ju, Gu, Mao, Liu, Yuan, and Zhang}]{luo2023self}
Luo, X.; Ju, W.; Gu, Y.; Mao, Z.; Liu, L.; Yuan, Y.; and Zhang, M. 2023.
\newblock Self-supervised graph-level representation learning with adversarial contrastive learning.
\newblock \emph{ACM Transactions on Knowledge Discovery from Data}, 18(2): 1--23.

\bibitem[{McInnes, Healy, and Melville(2018)}]{mcinnes2018umap}
McInnes, L.; Healy, J.; and Melville, J. 2018.
\newblock Umap: Uniform manifold approximation and projection for dimension reduction.
\newblock \emph{arXiv preprint arXiv:1802.03426}.

\bibitem[{Monti et~al.(2017)Monti, Boscaini, Masci, Rodola, Svoboda, and Bronstein}]{monti2017geometric}
Monti, F.; Boscaini, D.; Masci, J.; Rodola, E.; Svoboda, J.; and Bronstein, M.~M. 2017.
\newblock Geometric deep learning on graphs and manifolds using mixture model cnns.
\newblock In \emph{Proceedings of the IEEE conference on computer vision and pattern recognition}, 5115--5124.

\bibitem[{Morris et~al.(2020)Morris, Kriege, Bause, Kersting, Mutzel, and Neumann}]{morris2020tudataset}
Morris, C.; Kriege, N.~M.; Bause, F.; Kersting, K.; Mutzel, P.; and Neumann, M. 2020.
\newblock Tudataset: A collection of benchmark datasets for learning with graphs.
\newblock \emph{arXiv preprint arXiv:2007.08663}.

\bibitem[{Oord, Li, and Vinyals(2018)}]{oord2018representation}
Oord, A. v.~d.; Li, Y.; and Vinyals, O. 2018.
\newblock Representation learning with contrastive predictive coding.
\newblock \emph{arXiv preprint arXiv:1807.03748}.

\bibitem[{Paszke(2019)}]{paszke2019pytorch}
Paszke, A. 2019.
\newblock Pytorch: An imperative style, high-performance deep learning library.
\newblock \emph{arXiv preprint arXiv:1912.01703}.

\bibitem[{Radford et~al.(2021)Radford, Kim, Hallacy, Ramesh, Goh, Agarwal, Sastry, Askell, Mishkin, Clark et~al.}]{radford2021learning}
Radford, A.; Kim, J.~W.; Hallacy, C.; Ramesh, A.; Goh, G.; Agarwal, S.; Sastry, G.; Askell, A.; Mishkin, P.; Clark, J.; et~al. 2021.
\newblock Learning transferable visual models from natural language supervision.
\newblock In \emph{International conference on machine learning}, 8748--8763. PmLR.

\bibitem[{Rong et~al.(2019)Rong, Huang, Xu, and Huang}]{rong2019dropedge}
Rong, Y.; Huang, W.; Xu, T.; and Huang, J. 2019.
\newblock Dropedge: Towards deep graph convolutional networks on node classification.
\newblock \emph{arXiv preprint arXiv:1907.10903}.

\bibitem[{Sterling and Irwin(2015)}]{sterling2015zinc}
Sterling, T.; and Irwin, J.~J. 2015.
\newblock ZINC 15--ligand discovery for everyone.
\newblock \emph{Journal of chemical information and modeling}, 55(11): 2324--2337.

\bibitem[{Sun et~al.(2019)Sun, Hoffmann, Verma, and Tang}]{sun2019infograph}
Sun, F.-Y.; Hoffmann, J.; Verma, V.; and Tang, J. 2019.
\newblock Infograph: Unsupervised and semi-supervised graph-level representation learning via mutual information maximization.
\newblock \emph{arXiv preprint arXiv:1908.01000}.

\bibitem[{Suresh et~al.(2021)Suresh, Li, Hao, and Neville}]{suresh2021adversarial}
Suresh, S.; Li, P.; Hao, C.; and Neville, J. 2021.
\newblock Adversarial graph augmentation to improve graph contrastive learning.
\newblock \emph{Advances in Neural Information Processing Systems}, 34: 15920--15933.

\bibitem[{Tan et~al.(2024)Tan, Li, Jiang, Zhang, and Okumura}]{tan2024community}
Tan, S.; Li, D.; Jiang, R.; Zhang, Y.; and Okumura, M. 2024.
\newblock Community-invariant graph contrastive learning.
\newblock \emph{arXiv preprint arXiv:2405.01350}.

\bibitem[{Tucker(1966)}]{tucker1966some}
Tucker, L.~R. 1966.
\newblock Some mathematical notes on three-mode factor analysis.
\newblock \emph{Psychometrika}, 31(3): 279--311.

\bibitem[{Veli{\v{c}}kovi{\'c} et~al.(2018)Veli{\v{c}}kovi{\'c}, Fedus, Hamilton, Li{\`o}, Bengio, and Hjelm}]{velivckovic2018deep}
Veli{\v{c}}kovi{\'c}, P.; Fedus, W.; Hamilton, W.~L.; Li{\`o}, P.; Bengio, Y.; and Hjelm, R.~D. 2018.
\newblock Deep graph infomax.
\newblock \emph{arXiv preprint arXiv:1809.10341}.

\bibitem[{Wang et~al.(2024{\natexlab{a}})Wang, Wang, Meng, and Wang}]{wang2024afans}
Wang, S.; Wang, C.; Meng, P.; and Wang, Z. 2024{\natexlab{a}}.
\newblock AFANS: Augmentation-Free Graph Contrastive Learning with Adversarial Negative Sampling.
\newblock In \emph{International Conference on Intelligent Computing}, 376--387. Springer.

\bibitem[{Wang et~al.(2021)Wang, Min, Chen, and Wu}]{wang2021multi}
Wang, Y.; Min, Y.; Chen, X.; and Wu, J. 2021.
\newblock Multi-view graph contrastive representation learning for drug-drug interaction prediction.
\newblock In \emph{Proceedings of the web conference 2021}, 2921--2933.

\bibitem[{Wang et~al.(2023{\natexlab{a}})Wang, Hu, He, and Li}]{wang2023recognizing}
Wang, Z.; Hu, H.; He, C.; and Li, P. 2023{\natexlab{a}}.
\newblock Recognizing wafer map patterns using semi-supervised contrastive learning with optimized latent representation learning and data augmentation.
\newblock In \emph{2023 IEEE International Test Conference (ITC)}, 141--150. IEEE.

\bibitem[{Wang et~al.(2024{\natexlab{b}})Wang, Liu, Weston, Tian, and Li}]{wang2024learning}
Wang, Z.; Liu, L.; Weston, S. R.~F.; Tian, S.; and Li, P. 2024{\natexlab{b}}.
\newblock On learning discriminative features from synthesized data for self-supervised fine-grained visual recognition.
\newblock In \emph{European Conference on Computer Vision}, 101--117. Springer.

\bibitem[{Wang, Somayaji, and Li(2024)}]{wang2024learn}
Wang, Z.; Somayaji, K.~N.; and Li, P. 2024.
\newblock Learn-by-Compare: Analog Performance Prediction using Contrastive Regression with Design Knowledge.
\newblock In \emph{Proceedings of the 61st ACM/IEEE Design Automation Conference}, 1--6.

\bibitem[{Wang et~al.(2023{\natexlab{b}})Wang, Wang, Chen, Hu, and Li}]{wang2023contrastive}
Wang, Z.; Wang, Y.; Chen, Z.; Hu, H.; and Li, P. 2023{\natexlab{b}}.
\newblock Contrastive learning with consistent representations.
\newblock \emph{arXiv preprint arXiv:2302.01541}.

\bibitem[{Wu et~al.(2025)Wu, Zang, Zou, Luo, Bai, Xiang, Li, and Dong}]{cite-key}
Wu, Y.; Zang, Z.; Zou, X.; Luo, W.; Bai, N.; Xiang, Y.; Li, W.; and Dong, W. 2025.
\newblock Graph attention and Kolmogorov--Arnold network based smart grids intrusion detection.
\newblock \emph{Scientific Reports}, 15(1): 8648.

\bibitem[{Wu et~al.(2018)Wu, Ramsundar, Feinberg, Gomes, Geniesse, Pappu, Leswing, and Pande}]{wu2018moleculenet}
Wu, Z.; Ramsundar, B.; Feinberg, E.~N.; Gomes, J.; Geniesse, C.; Pappu, A.~S.; Leswing, K.; and Pande, V. 2018.
\newblock MoleculeNet: a benchmark for molecular machine learning.
\newblock \emph{Chemical science}, 9(2): 513--530.

\bibitem[{Xia et~al.(2021)Xia, Wu, Wang, Chen, and Li}]{xia2021progcl}
Xia, J.; Wu, L.; Wang, G.; Chen, J.; and Li, S.~Z. 2021.
\newblock Progcl: Rethinking hard negative mining in graph contrastive learning.
\newblock \emph{arXiv preprint arXiv:2110.02027}.

\bibitem[{Xu et~al.(2024)Xu, Chen, Li, Yang, Wang, Hu, and Ngai}]{xu2024fourierkangcffourierkolmogorovarnoldnetwork}
Xu, J.; Chen, Z.; Li, J.; Yang, S.; Wang, W.; Hu, X.; and Ngai, E. C.~H. 2024.
\newblock FourierKAN-GCF: Fourier Kolmogorov-Arnold Network -- An Effective and Efficient Feature Transformation for Graph Collaborative Filtering.
\newblock arXiv:2406.01034.

\bibitem[{You et~al.(2021)You, Chen, Shen, and Wang}]{you2021graph}
You, Y.; Chen, T.; Shen, Y.; and Wang, Z. 2021.
\newblock Graph contrastive learning automated.
\newblock In \emph{International conference on machine learning}, 12121--12132. PMLR.

\bibitem[{You et~al.(2020)You, Chen, Sui, Chen, Wang, and Shen}]{you2020graph}
You, Y.; Chen, T.; Sui, Y.; Chen, T.; Wang, Z.; and Shen, Y. 2020.
\newblock Graph contrastive learning with augmentations.
\newblock \emph{Advances in neural information processing systems}, 33: 5812--5823.

\bibitem[{Zhang et~al.(2024)Zhang, Fan, Liu, Huang, Zhao, Huang, and Liu}]{zhang2024expressive}
Zhang, B.; Fan, C.; Liu, S.; Huang, K.; Zhao, X.; Huang, J.; and Liu, Z. 2024.
\newblock The expressive power of graph neural networks: A survey.
\newblock \emph{IEEE Transactions on Knowledge and Data Engineering}.

\bibitem[{Zhang and Zhang(2024)}]{zhang2024graphkanenhancingfeatureextraction}
Zhang, F.; and Zhang, X. 2024.
\newblock GraphKAN: Enhancing Feature Extraction with Graph Kolmogorov Arnold Networks.
\newblock arXiv:2406.13597.

\bibitem[{Zhang, Yang, and Shi(2024)}]{zhang2024adaptive}
Zhang, Q.; Yang, C.; and Shi, C. 2024.
\newblock Adaptive negative representations for graph contrastive learning.
\newblock \emph{AI Open}, 5: 79--86.

\bibitem[{Zhu et~al.(2020)Zhu, Xu, Yu, Liu, Wu, and Wang}]{zhu2020deep}
Zhu, Y.; Xu, Y.; Yu, F.; Liu, Q.; Wu, S.; and Wang, L. 2020.
\newblock Deep graph contrastive representation learning.
\newblock \emph{arXiv preprint arXiv:2006.04131}.

\bibitem[{Zhu et~al.(2021)Zhu, Xu, Yu, Liu, Wu, and Wang}]{zhu2021graph}
Zhu, Y.; Xu, Y.; Yu, F.; Liu, Q.; Wu, S.; and Wang, L. 2021.
\newblock Graph contrastive learning with adaptive augmentation.
\newblock In \emph{Proceedings of the web conference 2021}, 2069--2080.

\end{thebibliography}

\newpage
\appendix

\begin{algorithm*}[htb]
   \caption{Higher-Order Singular Value Decomposition (Tucker Decomposition).}
   \label{alg:hosvd}
\begin{algorithmic}[1]
    \State {\bfseries Input:} An $I$-way tensor $\ten{C}$.
    \For{$i = 1$ to $I$}
    \State $\mat{C}^i$ = {\rm Unfold} ($\ten{C}, i$),
    {\bf \textit{\# Unfold the tensor along the $\rm i^{th}$ mode.}}
    \State $\mat{U}^{(i)}$, $\mat{\Sigma}^{(i)}$, $\mat{V}^{(i)}$ = {\rm SVD} ($\mat{C}^i$),
     {\bf \textit{\# Perform singular value decomposition on the reshaped tensor.}}
    \State $\mat{U}^{(i)}=\mat{U}^{(i)}[:,:r_i]$ {\bf \textit{\# Save the truncated singular vectors $\mat{U}^{(i)}$ as the model-i basis.}}
    \EndFor
    \State $\ten{G} = \ten{C}$, {\bf \textit{\# Initialize the tensor core with the $I$-way tensor $\ten{C}$.}}
    \For{$i = 1$ to $I$}
    \State $\ten{G}$ = $\ten{G} \times_i (\mat{U}^{(i)})^T$,
    {\bf \textit{\# Multiply the core tensor by the $\rm i^{th}$ orthogonal basis.}}
    \EndFor
    \State {\bfseries Output:} Core tensor $\ten{G}$, orthogonal basis $\mat{U}^{(1)}$, $\mat{U}^{(2)}$, $\cdots$, $\mat{U}^{(I)}$.
\end{algorithmic}
\end{algorithm*}

\begin{algorithm*}[htb]
   \caption{Algorithm flow of independent feature identification.}
   \label{alg:independent_feature}
\begin{algorithmic}[1]
    \State {\bfseries Input:} coefficient $\ten{C}\in \mathbb{R}^{d_{in}\times d_{out}\times d_{c}}$ of a KAN layer.
    \For{$i = 1$ to $d_c$}
    \State $\ten{C}^{(-j)}=\ten{C}_{:,\{1,\dots,d_{out}\}\setminus\{i\},:},$ {\bf \textit{\# $j_{th}$ mode-$2$ slice removed from $\ten{C}$.}}
    \State $\ten{G},\mat{U}^{(1)},\mat{U}^{(2)},\mat{U}^{(3)}=\mathrm{HOSVD}(\ten{C}^{(-j)})$ {\bf \textit{\# Call Algorithm~\ref{alg:hosvd} to decompose $\ten{C}^{(-j)}$.}}
    \State $\ten{P}=\ten{G}\times_1\mat{U}^{(1)}\times_2\mat{I}\times_3\mat{U}^{(3)}$
    {\bf \textit{\# $\ten{P}$ is a partially reconstructed $\ten{C}^{(-j)}$.}}
    \State $\tilde{\mat{M}}^{(2)}=\mathrm{arg}\min_{\tilde{\mat{M}}^{(2)}}\Vert\ten{C}_{:,i,:}-\sum_{s=1}^{r_2}\tilde{M}_s^{(2)}\ten{P}_{:,s,:}\Vert_F^2$
    {\bf \textit{\# Project the missing slice back to the basis to get $\tilde{\mat{M}}^{(2)}=\{\tilde{M}_s^{(2)}\}$ by solving the least-squares.}}
    \State $\tilde{\mat{U}}^{(2)}_i=[\mat{U}^{(2)}_{1},\dots,\mat{U}^{(2)}_{i-1},\tilde{\mat{M}}^{(2)},\mat{U}^{(2)}_{i+1}\dots]$ {\bf \textit{\# Plug $\tilde{\mat{M}}^{(2)}$ to $\mat{U}^{(2)}$.}}
    \State $\tilde{\ten{C}}_j=\ten{G}\times_1\mat{U}^{(1)}\times_2\tilde{\mat{U}}_j^{(2)}\times_3\mat{U}^{(3)}$   {\bf\textit{ \# Reconstruct $\ten{C}$.}}
    \State $\delta_j=\Vert\tilde{\ten{C}}_j-\ten{C}\Vert_F$   {\bf \textit{\# Reconstruction error.}}
    \EndFor
    \State {\bfseries Output:} reconstruction error $\bm{\delta}=\{\delta_i\}$.
\end{algorithmic}
\end{algorithm*}

\section{Proofs}

\subsection{Kolmogorov-Arnold Network Layer Output Dependency.}
We denote the coefficient of a KAN layer with $d_{in}$-dimensional input and $d_{out}$-dimensional output by $\ten{C}=\{c_{ijk}\}\in \mathbb{R}^{d_{in}\times d_{out}\times d_{c}}$. Here each B-spline curve is defined with $d_c$ coefficients. Given an input $\mat{x}\in\mathbb{R}^{d_{in}}$, the output $\mat{y}$'s $j$-th element can be written as:
\begin{equation}
    y_j = \sum_i\sum_k c_{ijk} B_{ijk}(x_i)
\end{equation}
$B_{ijk}(\cdot)$ is the $k$-th basis function of the edge that connects the $i_{th}$ input dimension to the $j_{th}$ output dimension.

We denote the slice corresponding to output dimension $d$ by $\ten{C}_{:,d,:}\in \mathbb{R}^{d_{in}\times d_{c}}$. If for some output index $d$, there exist indices $d_1,d_2,\dots,d_n\neq d$ and scalars $\{a_{d_l}\}_{l=1}^n$ such that 
\begin{equation}
    \ten{C}_{:,d,:}=\sum_{l=1}^n a_{d_l}\ten{C}_{:,d_l,:}\space,
\end{equation}
then the corresponding output feature $y_d$ of the KAN layer can be expressed as 
{\fontsize{8pt}{10pt}\selectfont
\begin{equation}
\begin{split}
    y_d &= \sum_i\sum_k c_{idk} B_{idk}(x_i)\\
    &=\sum_i\sum_k (a_{d_1}c_{id_1k}+a_{d_2}c_{id_2k}+\dots+a_{d_n}c_{id_nk}) B_{idk}(x_i)\\
    &=a_{d_1}y_{d_1}+a_{d_2}y_{d_2}+\dots+a_{d_n}y_{d_n},
\end{split}
\end{equation}}
which is the linear combination of the output features $y_{d_1},y_{d_2},\dots,y_{d_n}$.

In other words, any output dimension whose coefficient slice lies in the linear span of other slices will produce an output feature that is a linear combination of those corresponding dimensions. Therefore, we select those output dimensions whose corresponding coefficients can not be expressed as a linear combination of others as independent dimensions. These dimensions contain lower levels of noise and redundancy, making them especially informative for discriminative tasks.

\subsection{Variance Bound of Uniform B-spline Function}
We denote a uniform B-spline function $\phi(x)$ as:
\begin{equation}
    \phi(x)=\sum_kc_kB_k(x)
\end{equation}
Here $B_k(x)$ denotes the piecewise polynomial functions, and $c_k$ represents their corresponding coefficients. The mean of a B-spline function $\phi(x)$ is denoted by $\mu_{\phi}$, i.e.,
\begin{equation}
\begin{aligned}
    \mu_{\phi}&=\frac{1}{b-a}\int_a^b\phi(x)dx=\frac{1}{b-a}\int_a^b\sum_kc_kB_k(x)dx\\
    &=\frac{1}{b-a}\sum_kc_k\int_a^b B_k(x)dx \\
    &=\frac{1}{b-a}\sum_kc_k\omega_k
\end{aligned}
    \label{eqap2}
\end{equation}
Let $\omega_k$ denote $\int_a^b B_k(x)dx$. When using uniform grids in the domain $(a,b)$ of $\phi(x)$, all $\omega_k$ share the same value $\omega$ for any $k$. And as uniform B-spline basis functions satisfy:
\begin{equation}
    \sum_kB_k(x)=1 \qquad \forall x
    \label{eqap3}
\end{equation}
Integral both side of Equation~\ref{eqap3}:
\begin{equation}
    \sum_k\int_a^bB_k(x)dx=\int_a^b1dx=b-a
\end{equation}
For uniform B-spline, we thus have:
\begin{equation}
    \omega=\frac{b-a}{n}
    \label{eqap5}
\end{equation}
Here $n$ represent the number of coefficients. Plugging Equation~\ref{eqap5} into Equation~\ref{eqap2}, we get:
\begin{equation}
    \mu_{\phi}=\frac{1}{b-a}\sum_kc_k\omega_k=\frac{1}{n}\sum_kc_k=\bar c
\end{equation}
The mean of coefficients $\{c_k\}$ is denoted by $\bar c$.

Therefore, the variance of $\phi(x)$ can be written as:
{\fontsize{9pt}{10pt}\selectfont
\begin{equation}
    \begin{aligned}
        \mathrm{Var}[\phi(x)]&=\int(\phi(x)-\mu_{\phi})^2dx \\
        &=\int(\sum_kc_kB_k(x)-\bar c)^2dx \\
        &=\int(\sum_kc_kB_k(x)-\bar c\sum_kB_k(x))^2dx \\
        &=\int(\sum_k(c_k-\bar c)B_k(x))^2dx \\
        &=\int\sum_i\sum_j(c_i-\bar c)(c_j-\bar c)B_i(x)B_j(x)dx \\
        &=\sum_i\sum_j(c_i-\bar c)(c_j-\bar c)\int B_i(x)B_j(x)dx
    \end{aligned}
    \label{eqap7}
\end{equation}}
We use the following notation:
\begin{equation}
\begin{aligned}
    M_{ij}&=\int B_i(x)B_j(x)dx\\
    d_i&=c_i-\bar c
\end{aligned}
\end{equation}
Then Equation~\ref{eqap7} can be expressed as:
\begin{equation}
    \mathrm{Var}[\phi(x)]=\mathbf{d}^T\mathbf{M}\mathbf{d}
\end{equation}
where $\mathbf{M}=[M_{ij}]$ and $\mathbf{d}=[d_i]$ are a matrix and a vector, respectively.

Although the following derivation is applicable to B-spline functions of any order, as we used cubic (third-order) B-splines throughout our implementations, we present it for cubic B-splines. In the case of a uniform cubic B-spline, $B_k(x)$ overlaps non-zero with at most 3 neighbors on each side (indices $k\pm1$, $k\pm2$, $k\pm3$). As a result, $M_{ij} = 0$ for $|i-j|>3$. We can denote the distinct overlap integrals by $M(0)$ for $i=j$, $M(1)$ for $|i-j|=1$, $M(2)$ for $|i-j|=2$, and $M(3)$ for $|i-j|=3$ (with $M(k)=0$ for $k\ge4$). The matrix $\mathbf{M}$ can be written as:
{\fontsize{8pt}{10pt}\selectfont
\begin{equation}  
\mathbf{M} =
\begin{bmatrix}
M(0) & M(1) & M(2) & M(3) & 0     &\cdots & 0 \\
M(1) & M(0) & M(1) & M(2) & M(3) &\cdots & 0 \\
M(2) & M(1) & M(0) & M(1) & M(2) &\cdots & 0 \\
M(3) & M(2) & M(1) & M(0) & M(1) &\cdots & 0 \\
0    & M(3) & M(2) & M(1) & M(0) &\cdots & 0 \\
\vdots& \vdots& \vdots& \vdots&\vdots &\ddots & \vdots \\
0    & 0     & 0     & 0     & 0     &\cdots & M(0)
\end{bmatrix}
\end{equation}}

Then we can express $\mathrm{Var}[\phi(x)]$ as:
\begin{equation}
\begin{split}
    \mathrm{Var}[\phi(x)]&=M(0)\sum_id_i^2+2M(1)\sum_id_id_{i+1}\\
    &+2M(2)\sum_id_id_{i+2}+2M(3)\sum_id_id_{i+3}
\end{split}
\end{equation}

\begin{algorithm*}[htb]
   \caption{Algorithm flow of Khan-GCL (Pytorch-like style).}
   \label{alg:Khan-GCL}
\begin{algorithmic}[1]
    \State {\bfseries Input:} Initial KAN encoder parameters $\bm{\theta_e}$; Initial projection head parameters $\bm{\theta_p}$; Unlabeled dataloader; Hyperparameters $\epsilon_\delta$, $\epsilon_\rho$,$\sigma_\rho$,$\sigma_\delta$.
    \For{$\mathrm{x}^0$ in $\mathrm{dataloader}$}
    \State {\bf \textit{$^{\prime\prime\prime}$Representations of augmented graphs in a mini-batch$^{\prime\prime\prime}$}}
    \State $\mathrm{x}^\prime,\mathrm{x}^{\prime\prime}=\mathrm{A}_1(\mathrm{x}^0),\mathrm{A}_2(\mathrm{x}^0)$
    \State $\mathrm{z}^{\prime},\mathrm{z}^{\prime\prime}=\mathrm{encoder}(\mathrm{x}^\prime),\mathrm{encoder}(\mathrm{x}^{\prime\prime})$
    \State {\bf \textit{$^{\prime\prime\prime}$Identify critical dimensions and calculate perturbations$^{\prime\prime\prime}$}}
    \State $\mathrm{delta}=\mathrm{independent\_score}(\theta_e)$  {\bf\textit{\# See Algorithm~\ref{alg:independent_feature} and Section 4.2.1 for details.}}
    \State $\mathrm{rho}=\mathrm{discriminative\_score}(\theta_e)$  {\bf \textit{\# See Section 4.2.2 for details.}}
    \State $\mathrm{alpha}=\mathrm{rademacher}(0.5)$  {\bf \textit{\# rademacher distribution can be implemented by pytorch built-in}}
    \State $\qquad\qquad\qquad\qquad\qquad\quad$ {\bf function like torch.rand.}
    \State $\mathrm{p_\delta},\mathrm{p_\rho}=\mathrm{normal}(\mathrm{mean}=\epsilon_\delta*\mathrm{delta},\mathrm{std}=\sigma_\delta),\mathrm{normal}(\mathrm{mean}=\epsilon_\rho*\mathrm{rho},\mathrm{std}=\sigma_\rho)$  
    \State $\mathrm{z_{hard}}^\prime=\mathrm{z}^\prime+\mathrm{alpha}*(\mathrm{p_\rho}+\mathrm{p_\delta})$
    \State $\mathrm{alpha}=\mathrm{rademacher}(0.5)$  
    \State $\mathrm{p_\delta},\mathrm{p_\rho}=\mathrm{normal}(\mathrm{mean}=\epsilon_\delta*\mathrm{delta},\mathrm{std}=\sigma_\delta),\mathrm{normal}(\mathrm{mean}=\epsilon_\rho*\mathrm{rho},\mathrm{std}=\sigma_\rho)$
    \State $\mathrm{z_{hard}}^{\prime\prime}=\mathrm{z}^{\prime\prime}+\mathrm{alpha}*(\mathrm{p_\rho}+\mathrm{p_\delta})$
    \State {\bf \textit{$^{\prime\prime\prime}$Project to V space by projection head$^{\prime\prime\prime}$}}
    \State $\mathrm{v}^{\prime},\mathrm{v}^{\prime\prime},\mathrm{v_{hard}}^{\prime},\mathrm{v_{hard}}^{\prime\prime}=\mathrm{proj}(\mathrm{z}^\prime),\mathrm{proj}(\mathrm{z}^{\prime\prime}),\mathrm{proj}(\mathrm{z_{hard}}^\prime),\mathrm{proj}(\mathrm{z_{hard}}^{\prime\prime})$
    \State {\bf \textit{$^{\prime\prime\prime}$Calculate loss and optimize networks$^{\prime\prime\prime}$}}
    \State $\mathrm{Loss\_CL}=\mathrm{contrast\_loss}(\mathrm{v}^{\prime},\mathrm{v}^{\prime\prime})$
    \State $\mathrm{Loss\_HN}=\mathrm{hn\_loss}(\mathrm{v}^{\prime},\mathrm{v_{hard}}^{\prime}.\mathrm{detach()})+\mathrm{hn\_loss}(\mathrm{v}^{\prime\prime},\mathrm{v_{hard}}^{\prime\prime}.\mathrm{detach()})$
    \State $\mathrm{Loss\_Khan}=\mathrm{Loss\_CL}+\mathrm{Loss\_HN}$
    \State $\mathrm{Loss\_Khan}.\mathrm{backward()}$
    \State $\mathrm{update}(\bm{\theta_e})$
    \State $\mathrm{update}(\bm{\theta_p})$
    \EndFor
    \State {\bfseries Output:} Pre-trained encoder parameters $\bm{\theta_e}$.
\end{algorithmic}
\end{algorithm*}

As $M(k)\geq0$ for any $k\in\{0,1,2,3\}$, we have:

{\fontsize{9pt}{10pt}\selectfont
\begin{equation}
    \mathrm{Var}[\phi(x)]\leq M(0)\sum_id_i^2=M(0)\sum_i(c_i-\bar c)^2=M(0)\sigma_c^2
\end{equation}}

\section{Higher-order Singular Value Decomposition}
In Algorothm~\ref{alg:hosvd}, we present the detailed process of Higher-Order Singular Value Decomposition (HOSVD), also known as Tucker Decomposition~\cite{tucker1966some,de2000multilinear}. For each 3-way coefficient tensor $\ten{C} \in \mathbb{R}^{n_{in} \times n_{out} \times n_c}$ within the KAN layer architecture, HOSVD produces a decomposition consisting of:

\begin{enumerate}[%
  label=\roman*.,       
  labelwidth=1em,        
  labelsep=0.5em,          
  leftmargin=3em           
]
    \item A core tensor $\ten{G} \in \mathbb{R}^{r_1 \times r_2 \times r_3}$, and
    \item Three orthogonal basis matrices $\mathbf{U}^{(1)} \in \mathbb{R}^{n_{in} \times r_1}$, $\mathbf{U}^{(2)} \in \mathbb{R}^{n_{out} \times r_2}$, and $\mathbf{U}^{(3)} \in \mathbb{R}^{n_c \times r_3}$.
\end{enumerate}

Each factor matrix $\mathbf{U}^{(k)}$ contains the leading $r_k$ left singular vectors obtained from the SVD of the mode-$k$ unfolding of tensor $\mathcal{C}$. These truncated orthonormal bases preserve the most significant components of the original tensor while enabling substantial parameter reduction.

\section{Algorithm of Independent Feature Identification}
In Algorithm~\ref{alg:independent_feature}, we present the detailed algorithm of our proposed independent feature identification technique in Critical KAN Feature Identification (\textbf{CKFI}).

\section{Algorithm of Khan-GCL}
In Algorithm~\ref{alg:Khan-GCL}, we conclude the algorithm flow of our proposed Khan-GCL pre-training approach in Pytorch-like~\cite{paszke2019pytorch,fey2019fast} style.

\section{Datasets and Evaluation Protocols}
\subsection{Datasets}
We summarize the characteristics of all datasets utilized in our experiments in Table~\ref{tab:datasets_details}. For transfer learning experiments, we pre-train our encoder on Zinc-2M~\cite{sterling2015zinc} and evaluate its performance across 8 datasets from MoleculeNet~\cite{wu2018moleculenet}: BBBP, Tox21, SIDER, ClinTox, MUV, HIV, and BACE. In our unsupervised learning evaluations, we assess our method on 8 datasets from TU-datasets~\cite{morris2020tudataset}, comprising NCI1, PROTEINS, DD, MUTAG, COLLAB, RDT-B, RDT-M5K, and IMDB-B. Additionally, we use MNIST-superpixel~\cite{monti2017geometric} in our experiments.

\begin{table*}
\small
  \centering
  \begin{tabular}{c|c|c|c|c}
    \toprule
    Datasets & Domain & Dataset size & Avg. node per graph & Avg. degree \\
    \midrule
    Zinc-2M   & Biochemical & 2,000,000 & 26.62 & 57.72 \\
    BBBP      & Biochemical & 2,039     & 24.06 & 51.90 \\
    Tox21     & Biochemical & 7,831     & 18.57 & 38.58 \\
    SIDER     & Biochemical & 1,427     & 33.64 & 70.71 \\
    ClinTox   & Biochemical & 1,477     & 26.15 & 55.76 \\
    MUV       & Biochemical & 93,087    & 24.23 & 52.55 \\
    HIV       & Biochemical & 41,127    & 25.51 & 54.93 \\
    BACE      & Biochemical & 1,513     & 34.08 & 73.71 \\
    \midrule
    NCI1      & Biochemical & 4,110     & 29.87 & 1.08  \\
    PROTEINS  & Biochemical & 1,113     & 39.06 & 1.86  \\
    DD        & Biochemical & 1,178     & 284.32 & 715.66  \\
    MUTAG     & Biochemical & 188       & 17.93 & 19.79  \\
    COLLAB    & Social Networks & 5,000 & 74.49 & 32.99  \\
    RDT-B     & Social Networks & 2,000 & 429.63 & 1.16  \\
    RDT-M5K   & Social Networks & 4,999 & 508.52 & 1.17  \\
    IMDB-B    & Social Networks & 1,000 & 19.77 & 96.53  \\
    \midrule
    MNIST-superpixel & Superpixel & 70,000 & 70.57 & 8 \\
    \bottomrule
      \end{tabular}
\caption{Details of datasets used in our experiments.}
  \label{tab:datasets_details}
\end{table*}

\begin{table*}[htb]
 \small
  \centering
  \begin{tabular}{c|c|c|c}
    \toprule
    Experiments & Transfer Learning & Unsupervised Learning & MNIST-Superpixel \\
    \midrule
    GNN Type                   & GIN & GIN & GIN \\
    Encoder Neuron Number         & [300,300,300,300,300] & [32,32,32] & [110,110,110,110] \\
    Projection Head Neuron Number & [300,300] & [32,32]     & [110,110] \\
    Pooling Layer              & Global Mean Pool & Global Add Pool & Global Add Pool \\
    \bottomrule
      \end{tabular}
\caption{Details of Model Architecture.}
  \label{tab:model_arch}
\end{table*}

\subsection{Evaluation Protocols}
In this section, we present the tasks for evaluating our self-supervised learning framework. Detailed hyperparameters of these tasks are provided in Appendix~\ref{sec:model_arch_and_hyperparam}.
\paragraph{Transfer learning.}
First, we pre-train the encoder on a large biochemical dataset Zinc-2M~\cite{sterling2015zinc} using our proposed Khan-GCL framework. After pre-training, we discard the projection head and append a linear layer after the pre-trained encoder. Then we perform end-to-end fine-tuning of the encoder and linear layer on the training sets of the downstream datasets. Test ROC-AUC at the best validation epoch is reported. Each downstream experiment is performed 10 times, and the mean and standard deviation of ROC-AUC are reported. We refer to \cite{hu2019strategies} and \cite{you2020graph} for more details of this task. 

\paragraph{Unsupervised learning.}
In this task, we follow the setup of \cite{sun2019infograph}. Specifically, we pre-train the encoder on TU-datasets~\cite{morris2020tudataset}. While evaluating the encoder, we use a SVM classifier to classify the output representations of the encoder. We report 10-fold cross validation accuracy averaged for 5 runs. 

\paragraph{More configurations of Khan-GCL in unsupervised learning.}
In Section 5.2, to perform ablation study regarding CKFI and the hard negative generation approach, we introduce three additional configurations of Khan-GCL, i.e., `w/ random-perturb', `w/o d-dims', and `w/o i-dims'. Here we provide details of these three configurations.

\begin{enumerate}[%
  label=\roman*.,       
  labelwidth=1em,        
  labelsep=0.5em,          
  leftmargin=3em           
]
    \item `w/ random-perturb': perturbations are sampled from random Gaussian distribution, i.e., for a graph representation $\mat{z}_j$, $\mat{z}_j^{hard}=\mat{z}_j+\mat{p}^{rand}$, where $\mat{p}^{rand}$ is sampled as follows:
    
    \begin{equation}
        \mathbf{p}^{rand}=\{p_i^{rand}=\alpha_i\cdot u_i^{rand}\}
        \label{eq:random_perturb}
    \end{equation}
    in which:
    \begin{equation}
    u_i^{rand}\sim\mathcal{N}(\epsilon_{rand}\sigma^2_{rand}),\quad\alpha_i\sim \mathrm{Rad}
    \end{equation}
    \item `w/o d-dims': perturbations are applied only to the independent dimensions, i.e., for a graph representation $\mat{z}_j$, $\mat{z}_j^{hard}=\mat{z}_j+\mat{p}^{\delta}$
    \item `w/o i-dims': perturbations are applied only to the discriminative dimensions, i.e., for a graph representation $\mat{z}_j$, $\mat{z}_j^{hard}=\mat{z}_j+\mat{p}^{\rho}$
\end{enumerate}

\paragraph{MNIST-superpixel.} We follow the settings of \cite{you2020graph} in pre-training the encoder on MNIST-superpixel dataset~\cite{monti2017geometric}. 

While performing nearest neighbor retrieval, for a certain graph $\mat{x}_j$ (whose representation is $\mat{z}_j$), we first generate a hard negative $\mat{z}_j^{hard}$ of it by applying the perturbations. Then we search the representations of all graphs across the dataset to find the 5 graphs with the largest latent similarity (measured by cosine similarity) with $\mat{z}_j^{hard}$.

\begin{figure*}[htb]
  \centering
  \includegraphics[width=\textwidth]{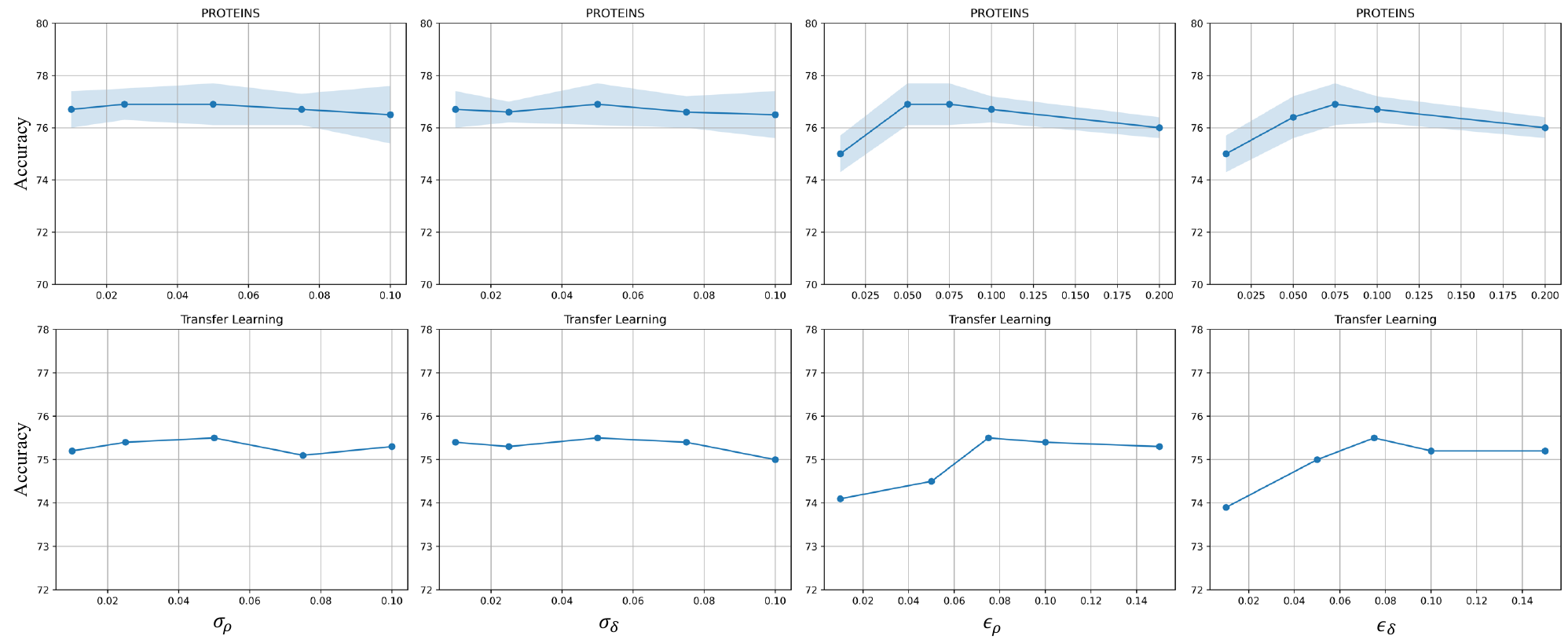}
  \caption{Ablation study results regarding hyperparameters.}
  \label{fig:hyper_ablation}
\end{figure*}

\section{Detailed Experiment Settings}
\label{sec:model_arch_and_hyperparam}
\subsection{Encoder Architecture}
For the sake of fairness across all comparison against existing methods, we follow \cite{you2020graph} for the input, output, and hidden layer size of the encoder. Details of the encoder are concluded in Table~\ref{tab:model_arch}.

In Khan-GCL, we replace all MLPs in the encoder by same-sized KANs. In all implementations, we use cubic B-spline-based KANs where the grid sizes are set to 5.

\subsection{Transfer Learning Settings}
In transfer learning pre-training, we train the encoder using an Adam optimizer~\cite{kingma2014adam} with initial learning rate $1\times10^{-4}$ for 100 epochs. The temperature hyperparameter $\tau$ in contrastive loss is $0.1$. In hard negative generation, we choose $\epsilon_\delta=\epsilon_\rho=0.075$ and $\sigma_\delta=\sigma_\rho=0.05$.

\subsection{Unsupervised Learning Settings}
In unsupervised learning pre-training, the encoder is optimized by an Adam optimizer~\cite{kingma2014adam} with initial learning rate $1\times10^{-4}$ for 60 epochs. The temperature hyperparameter $\tau$ in contrastive loss is $0.2$. While generating perturbations, we choose $\epsilon_\delta=\epsilon_\rho=0.075$ and $\sigma_\delta=\sigma_\rho=0.05$. 

\begin{table}[ht]
  \small
  \centering
  \setlength{\tabcolsep}{0.5mm}{
  \begin{tabular}{c|c|c|c}
    \toprule
    Methods   & GraphCL & RGCL & Khan-GCL(Ours) \\
    \midrule
    Time (second/iteration) & 0.046 & 0.059 & 0.063\\
    \bottomrule
      \end{tabular}}
\caption{Running time comparison in Zinc-2M pre-training.}
\label{tab:run_time}
\end{table}

\section{Training Time Comparison of KAN and MLP-based Encoder}
All experiments are conducted on a single NVIDIA A100 GPU. Table~\ref{tab:run_time} compares the runtime of Khan-GCL with several state-of-the-art GCL approaches (GraphCL~\cite{you2020graph}, RGCL~\cite{RGCL}) in Zinc-2M pre-training. Although the iterative computation involved in KAN's B-spline functions incurs additional training overhead, Khan-GCL achieves a runtime comparable to recent state-of-the-art GCL approaches.

\section{Additional Experiment Results}
\subsection{MNIST-superpixel classification}
In addition to the nearest neighbor retrieval experiment, we also present the classification accuracy of the pre-trained encoder on MNIST-superpixel in Table~\ref{tab:mnist_classify}. After pre-training, the encoder is fine-tuned on 1\% labeled data randomly sampled from the whole dataset.

\begin{table}[htb]
  \centering
  \begin{tabular}{c|c}
    \toprule
    Methods   & Acc (in \%) \\
    \midrule
    Infomax~\cite{sun2019infograph} & 63.2$\pm$0.8 \\
    GraphCL~\cite{you2020graph} & 83.4$\pm$0.3 \\
    RGCL~\cite{RGCL} & 83.8$\pm$0.4 \\
    Khan-GCL(Ours) & \textbf{84.2$\pm$0.3} \\
    \bottomrule
      \end{tabular}
      \caption{Performance comparison in MNIST-superpixel classification.}
  \label{tab:mnist_classify}
\end{table}

\subsection{Additional ablation study}
In this section, we perform additional ablation studies on the hyperparameters $\epsilon_\delta$, $\epsilon_\rho$, $\sigma_\delta$, and $\sigma_\rho$ of Khan-GCL. Figure~\ref{fig:hyper_ablation} presents the results of unsupervised learning on the PROTEINS dataset and averaged transfer learning performance with varying hyperparameter settings.

\section{Potential Societal Impacts}
By integrating expressive and interpretable Kolmogorov–Arnold Networks (KANs), Khan-GCL establishes a new state-of-the-art approach for self-supervised graph learning. This method is applicable to various real-world domains, including recommendation systems, cybersecurity, and drug discovery. We anticipate that leveraging KANs within GCL frameworks, along with our proposed feature identification techniques, will inspire further research in representation learning and self-supervised graph learning.

Nevertheless, since KAN architectures rely on iterative B-spline computations, they require more training time and computational resources compared to conventional MLPs. Consequently, this increased resource consumption may lead to environmental concerns, such as higher carbon emissions.

\section{Declaration of LLM usage}
We used LLMs only for writing assistance, e.g., grammar and spell checking, and did not rely on them for generating or analyzing core research content.

\end{document}